\pdfoutput=1

\documentclass[11pt]{article}

\usepackage[preprint]{acl}

\usepackage{times}
\usepackage{latexsym}
\usepackage[T1]{fontenc}
\usepackage[utf8]{inputenc}
\usepackage{microtype}
\usepackage{inconsolata}
\usepackage{graphicx}
\usepackage{xspace}
\usepackage{tabularx}
\usepackage{multirow}
\usepackage{graphicx}
\usepackage{amssymb}
\usepackage{pifont}
\usepackage{float}
\usepackage{enumitem}
\usepackage{amsmath}
\usepackage{times}
\usepackage{latexsym}
\usepackage{booktabs}
\usepackage{arydshln}

\newcommand{\ours}{DEP\xspace}

\newcommand{\eg}{\emph{e.g., }}

\newcommand{\cmark}{\textcolor{green}{\checkmark}}
\newcommand{\xmark}{\textcolor{red}{\ding{55}}}

\title{Latent Inter-User Difference Modeling for LLM Personalization}

\author{
 \textbf{Yilun Qiu\textsuperscript{1}},
 \textbf{Tianhao Shi\textsuperscript{2}},
 \textbf{Xiaoyan Zhao\textsuperscript{3}\footnotemark[1]},
\\
 \textbf{Fengbin Zhu\textsuperscript{1}},
 \textbf{Yang Zhang\textsuperscript{1}\thanks{Corresponding Authors}},
 \textbf{Fuli Feng\textsuperscript{2}}
\\
 \textsuperscript{1}National University of Singapore \\
  \textsuperscript{2}University of Science and Technology of China \\
 \textsuperscript{3}The Chinese University of Hong Kong \\
 \small
    qiuyilun@u.nus.edu, 
    sth@mail.ustc.edu.cn,
    xzhao@se.cuhk.edu.hk, 
 \\
 \small
    zhfengbin@gmail.com,
    zyang1580@gmail.com,
    fulifeng93@gmail.com
}

\begin{document}
\maketitle

\begin{abstract}

Large language models (LLMs) are increasingly integrated into users’ daily lives, leading to a growing demand for personalized outputs.
Previous work focuses on leveraging a user's own history, overlooking inter-user differences that are crucial for effective personalization.
While recent work has attempted to model such differences, the reliance on language-based prompts often hampers the effective extraction of meaningful distinctions.
To address these issues, we propose \textit{Difference-aware Embedding-based Personalization} (\ours), a framework that models inter-user differences in the latent space instead of relying on language prompts. 
\ours constructs soft prompts by contrasting a user’s embedding with those of peers who engaged with similar content, highlighting relative behavioral signals.
A sparse autoencoder then filters and compresses both user-specific and difference-aware embeddings, preserving only task-relevant features before injecting them into a frozen LLM.
Experiments on personalized review generation show that \ours consistently outperforms baseline methods across multiple metrics.
Our code is available at \url{https://github.com/SnowCharmQ/DEP}.


\end{abstract}

\section{Introduction}

With continuous advancements in general-purpose intelligence, large language models (LLMs)~\cite{gpt4,llama3,qwen3,gemma3,deepseekr1} are increasingly integrated into everyday life, assisting users in making decisions~\cite{react,zhao2025exploring}, retrieving information~\cite{zhao2024comprehensive,attentionrag}, and task management~\cite{mome,taskbench}.
This growing presence has raised expectations for LLMs to go beyond generic, one-size-fits-all responses and instead produce responses that align with individual users' unique preferences.
To meet these heightened expectations, there has been a growing interest in \textit{LLM personalization}
~\cite{prism,personalizationsurvey1,personalizationsurvey2,personalizationsurvey3,personalizationsurvey4}, which aims at tailoring model outputs based on user-specific information.

Most existing methods adopt the memory-retrieval paradigm~\cite{lamp,summ}, where user history is stored in memory, and key information is then retrieved as a steering prompt to guide model generation. Earlier works~\cite{teachllm,pearl} focused solely on retrieving information about the user themselves for personalization. However, recent work such as DPL~\cite{dpl} argues that effective personalization should also capture how a user differs from others. This view is grounded in insights from psychology and behavioral science~\cite{diff1,diff2,diff3}, which highlight that inter-user variability determines individuality and shapes users’ distinct preferences. Accordingly, DPL incorporates inter-user comparison in the retrieval history, formulating the comparison as a natural language inference task performed by the LLM.

Despite DPL’s demonstrated effectiveness, we argue that its language-based inter-user comparison paradigm using LLMs is structurally ill-suited for accurately extracting inter-user differences. On one hand, controlling the extraction of differences using an LLM is challenging; although providing extraction criteria can help, some aspects of distinction may be missed due to the difficulty of defining comprehensive standards. On the other hand, including other users’ raw data for comparison in LLMs can result in verbose prompts that strain the model’s context window, ultimately hindering the extraction of meaningful inter-user differences.

To address these limitations, we propose shifting to latent-space difference modeling, where task-relevant differences between users are structurally represented and compared in the latent embedding space~\cite{uem,pplug,commer,userllm}. Compared to natural language, latent embeddings offer two key advantages: (1) they encode fine-grained, context-dependent behavioral patterns in a compact form; and (2) they inherently support inter-user comparison through vector operations, enabling direct integration of comparison signals. Together, these properties make latent embeddings a more suitable medium for modeling inter-user differences within LLMs.
Building on this idea, we propose a new method called \textit{Difference-aware Embedding-based Personalization} (DEP), which models task-relevant inter-user differences in the latent space and injects them into LLMs as soft prompts for personalization. DEP extracts a difference-aware embedding as a soft prompt by subtracting and aggregating the user’s embedding against those of other users who engaged with similar items. At the same time, the original user-specific embedding is provided as a reference to supply contextual information.
Both embeddings are essential: the user-specific embedding defines the behavioral context, while the difference-aware embedding captures deviations from that context. Together, they form a contextualized inter-user signal that reflects both individualized preferences and relative differences.


Taking a step further, latent differences can be redundant, as not all aspects are task-relevant—some may simply introduce noise. To extract essential information while filtering out irrelevant signals, we process both user-specific and difference-aware embeddings using a sparse autoencoder (SAE)~\cite{sae}, which enforces sparsity to retain only key features. The resulting compressed representations are then injected into a frozen LLM as soft prompts. The SAE is fine-tuned to align these representations with the LLM’s internal understanding, allowing the model to effectively leverage inter-user differences for improved personalization.
We conduct extensive experiments on a representative task, review generation~\cite{amazon2018}, where \ours achieves state-of-the-art performance across multiple evaluation metrics.

Our main contributions are as follows:
\begin{itemize}[leftmargin=*]
\setlength{\itemsep}{0pt}
    \item 
    We propose modeling inter-user differences in the latent space to enable more comprehensive and flexible extraction of preference signals for LLM personalization.
    
    
    \item 
    We introduce a novel method, \ours, to achieve latent inter-user difference modeling, equipped with a sparse autoencoder to extract task-relevant differences while filtering out noise.
    
    \item Extensive experiments show that our \ours consistently outperforms baseline methods with significant improvements.
\end{itemize}

\section{Preliminary}

\noindent
\textbf{Problem Formulation.}
This work studies the task of LLM personalization, where the goal is to produce user-aligned output that reflects the individual preferences of a given user.
We assume that each user has a set of historical texts. These historical texts are utilized to help the LLM infer the user’s interests and generate personalized content.
Formally, let $D$ denote the collection of historical records from all users. Each record in $D$ is represented as $(u, i, y_{u}^i)$, where $u$ is a user, $i$ is an item (or object) the user has focused on, and $y_{u}^i$ denotes the text written or preferred by user $u$ for item $i$. 
When the target user $u^{\prime}$ submits a request to generate text for a target item $i^{\prime}$, the LLM is expected to produce an output that aligns with the preference of $u^{\prime}$ based on $D$.

Without loss of generality, this work focuses on review generation, a representative personalization task.
The goal is to generate reviews tailored to a user’s style and preferences, ensuring the output aligns with how the user typically expresses opinions on items such as movies or products.



\vspace{3pt}

\begin{figure*}[t]
    \centering
    \includegraphics[width=0.89\linewidth]{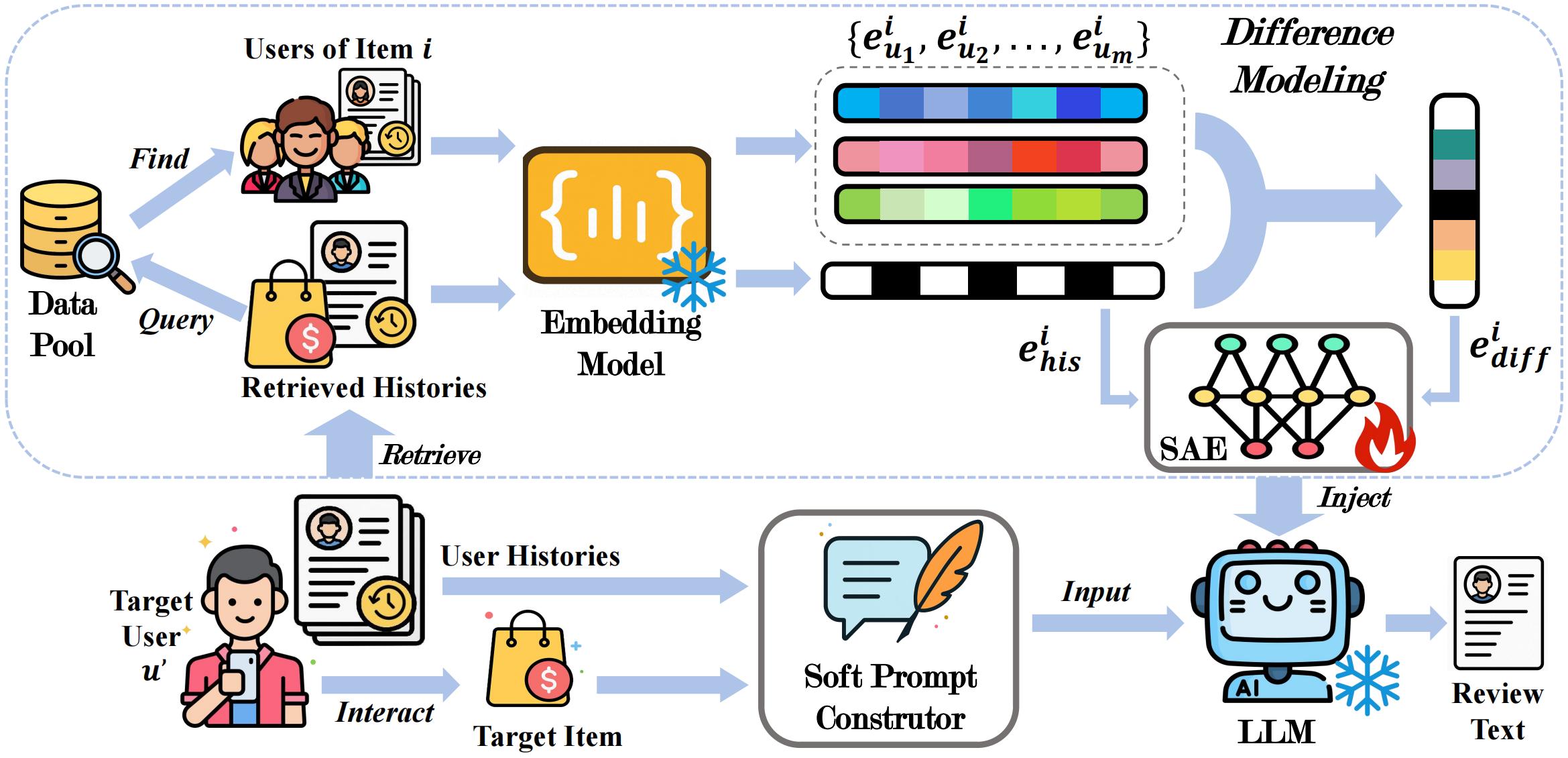}
    \caption{Overview of the proposed \ours method, which introduces user-specific and difference-aware embeddings to capture both individual preferences and inter-user differences. A sparse autoencoder (SAE) refines these representations, which are then injected into a frozen LLM as soft prompts to guide personalized text generation.}
    \label{main_method}
\end{figure*}

\noindent\textbf{Memory-retrieval framework.}
A common approach to enabling LLMs to perform personalized generation is to store users’ history and retrieve relevant signals at inference.
Following DPL~\cite{dpl}, effective personalization should capture both a user’s own behavioral patterns and how they differ from others.
This involves extracting key preference signals from two sources: the user’s own history, which reflects individual tendencies, and other users’ behaviors, which provide materials for modeling inter-user differences.
Formally, given a target user $u^\prime$ and a target item $i^\prime$, the personalized generation process can be formulated as:
\begin{equation}
    \hat{y}_{u^{\prime}}^{i^{\prime}} = \mathrm{LLM}(u^\prime, i^\prime, \phi({D_{u^{\prime}}; D})),
\end{equation}
where $\hat{y}_{u^\prime}^{i^\prime}$ denotes the generated text, $D_{u^{\prime}}$ denotes the history of the target user $u^\prime$, and $\phi({D_{u^{\prime}}; D})$ denotes the process that extracts user-specific and difference-aware preference signals from $D_u$ and $D$.
This memory retrieval framework supports lifelong user modeling without requiring LLM retraining, making it both adaptable and resource-efficient for real-world personalization scenarios.

\section{Methodology}


This section introduces our proposed \textit{Difference-aware Embedding-based Personalization} (\textbf{\ours}).
We begin with its motivation and an overview of the framework, followed by detailed descriptions of each key steps.

\subsection{Overview}

Personalization modeling requires capturing not only a user’s own behavioral patterns, but also how this user differs from others.
In modeling inter-user differences, existing work~\cite{dpl} relies on LLMs to summarize inter-user comparisons in natural language, which may miss some key aspects of distinctions during the summarization.
To address this limitation, we propose the DEP method, which aims to model inter-user differences in the latent space. 
DEP has three main parts: (1) constructing two representations to capture difference-aware preference: a user-specific embedding to model the behavioral context, and a difference-aware embedding to model how the user deviates from others within that context; 
(2) distilling the representations with a sparse autoencoder to retain informative preference signals; and
(3) injecting the compressed representation into a frozen LLM as soft prompts and fine-tuning the autoencoder to align this representation with the LLM's internal understanding. Figure~\ref{main_method} provides an overview of our proposed \ours. Next, we elaborate the three parts in detail.

\subsection{Difference-aware Embedding-based Personalization (\ours)}
In this section, we introduce three key steps of DEP: constructing latent difference-aware representations, distilling them via a sparse autoencoder, and injecting them into an LLM for personalization.


\subsubsection{Latent-space Difference-aware Representation Modeling}

The core of DEP is to model inter-user differences in the latent space 
through contrastive signals grounded in shared item contexts.
To achieve this, following the memory-retrieval paradigm~\cite{lamp,longlamp,dpl}, DEP first retrieves a set of representative interactions from the user’s history, which serve as anchors for inter-user comparison.
For a given user $u^\prime$, we assume a subset of $N$ key interactions, denoted as $D_{u^\prime}^{*}$, can be obtained via
retrieval~\cite{personalizationsurvey2} from $D_{u^\prime}$.
Then, for each retrieved interaction $(u^\prime, i, y_{u'}^i) \in D_{u^\prime}^{*}$, we aim to compare it with reviews written by other users for the same item $i$, which provides a natural basis for inter-user comparison.
To this end, we first encode the user's own review $y^i_{u'}$ using a frozen text embedding model $f_{\text{emb}}(\cdot)$ to obtain the user-specific embedding:
\begin{equation}\small
    e_\text{his}^{i}=f_{\text{emb}}(y_{u'}^i), 
\end{equation}
where $e_{his}^{i}$ denotes the user-specific embedding that reflects the preference pattern of 
user $u^\prime$ on item $i$. 
Next, to construct inter-user embeddings, we identify the set of peer users who also interacted with item $i$, excluding $u'$, as $\{u_{1}, u_{2}, \dots, u_{m}\}$, where $u_{j}$ denotes the $j$-th peer user of item $i$. Each peer user $u_j$ provides a review $y^i_{u_j}$, which is encoded into an embedding:
\begin{equation}\small
    e^{i}_{u_{j}}=f_{\text{emb}}(y^{i}_{u_{j}})).
\end{equation}
Then we compute the difference-aware embedding by aggregating the vector differences between the target user and each peer:
\begin{equation}\small
    e_\text{diff}^{i} = \frac{1}{m}\sum_{j=1}^m  (e_\text{his}^i - e_{u_j}^i), 
\end{equation}
where $e_\text{diff}^{i}$ denotes the difference-aware embedding. These two embeddings capture complementary perspectives: the user-specific embedding $e^i_\text{his}$ represents the behavior pattern of the target user and serves as a reference of context, while the difference-aware embedding $e^i_\text{diff}$ models how this behavior pattern relatively deviates from others under the context. Together, they form a structured representative to capture the inter-user differences.

\subsubsection{Sparse Representation Distillation}

While the user-specific and difference-aware embeddings capture rich semantic and contrastive signals, they may contain redundant or irrelevant information that hinders efficient personalization.
To address this, we apply a sparse autoencoder (SAE)~\cite{sae} to compress the high-dimensional embeddings into informative representations. The SAE adopts an encoder-decoder architecture with an $\ell_1$-based sparsity constraint on the latent space, encouraging the model to retain only the most salient features.
Given a history embedding $e^i_\text{his}$ and a difference-aware embedding $e^i_\text{diff}$, the encoder produces their respective low-dimensional latent vectors, $z^i_\text{his}$ and $z^i_\text{diff}$, formally:
\begin{equation}
\begin{aligned}
    z^i_\text{his} &= f_\text{enc}(e^i_\text{his}), \quad & \hat{e}^i_\text{his} &= f_\text{dec}(z^i_\text{his}), \\
    z^i_\text{diff} &= f_\text{enc}(e^i_\text{diff}), \quad & \hat{e}^i_\text{diff} &= f_\text{dec}(z^i_\text{diff}),
\end{aligned}
\end{equation}
where $f_\text{enc}(\cdot)$ and $f_\text{dec}(\cdot)$ denote the encoder and decoder networks, respectively. 
The encoder outputs $z^i_\text{his}$ and $z^i_\text{diff}$ are used as sparse preference representations for downstream soft prompt construction.


\subsubsection{Representation Injection}

After obtaining the distilled latent representations from the sparse autoencoder, we aim to integrate personalized signals into the generation process of a frozen LLM.
To achieve this, we adopt a soft prompt injection mechanism, where the compressed user-specific and difference-aware embeddings are projected into the input space of the LLM and used to condition its output without updating model parameters.

\noindent \textbf{Soft Prompt Construction and Injection.} 
For each retrieved history $(u', i, y)$, we obtain $z^i_\text{his}$ and $z^i_\text{diff}$ from the SAE encoder, corresponding to the user-specific and difference-aware embeddings.
These representations are projected into the LLM input space via a lightweight projection network $\mathcal{M}_p(\cdot)$, which aligns their dimensionality with that of the LLM's embedding layer:

\begin{equation}\small
    p^i_\text{his} = \mathcal{M}_p(z^i_\text{his}), \quad
p^i_\text{diff} = \mathcal{M}_p(z^i_\text{diff}), 
\end{equation}
where $p^i_\text{his}$ and $p^i_\text{diff}$ are resulting soft prompt vectors, which are injected into the input sequence at designated positions. Then, the personalized generation process given the target user $u^\prime$ and the target item $i^\prime$ is performed as:
\begin{equation}\small
\hat{y}_{u'}^{i'} = LLM\left(\mathcal{S}(i', \{i, p^i_\text{his}, p^i_\text{diff}\}_{{i}\in{I^*_{u'}}})\right),
\end{equation}
where $I^*_{u'}$ denotes the top-$N$ retrieved items from the target user's interacted history, and $\mathcal{S}(i', \{i, p^i_\text{his}, p^i_\text{diff}\}_{{i}\in{I^*_{u'}}})$ is a textual prompt constructed from both the target item $i'$ and the soft prompts to model inter-user differences, and the original user’s original review history to model user’s own writing patterns.
The template can be found in Figure~\ref{prompt} in Appendix~\ref{ov_prompt}.

\noindent \textbf{Training Objectives.}
To guide the SAE learning informative representation for LLM personalization and make the soft prompts align with the LLM's internal understanding, we jointly optimize two components: the SAE for latent representation learning and the projection network that maps its output into the LLM's input space for personalized generation.
Specifically, we employ a standard generation loss $\mathcal{L}_{\text{gen}}$, computed by the frozen LLM based on its generated output and the ground-truth personalized text, to supervise the training of the SAE and the projection network.
The SAE is trained with two standard objectives: a reconstruction loss to ensure information preservation, and a sparsity loss to promote selective preference encoding.
For the reconstruction loss, we adopt the Smooth L1 loss, which is formulated as follows:
\begin{equation}
\small
    \mathcal{L}_\text{recon} = \text{SmoothL1}(e^i_\text{his}, \hat{e}^i_\text{his}) + \text{SmoothL1}(e^i_\text{diff}, \hat{e}^i_\text{diff}).
\end{equation} 
The sparsity loss is applied to the distilled latent vector $z^i_\text{his} \in \mathbb{R}^{d'}$ and $ z^i_\text{diff} \in \mathbb{R}^{d'}$, encouraging the preservation of the most informative signals.
For each, we compute the average activation $\hat{\rho}_\text{his}$ and $\hat{\rho}_\text{diff}$ as:
\begin{equation}\small
    \hat{\rho}_\text{his} = \frac{1}{N} \sum_{i=1}^{N} z^i_\text{his}, \quad 
    \hat{\rho}_\text{diff} = \frac{1}{N} \sum_{i=1}^{N} z^i_\text{diff}.
\end{equation}
We then compute the sparsity loss by applying KL divergence between each of $\hat{\rho}_\text{his}$ and $\hat{\rho}_\text{diff}$ and a predefined sparsity target $\rho$.
\begin{equation}
    \small
    \mathcal{L}_\text{sparse} = \frac{1}{d'}\sum_{j=1}^{d'}KL(\rho||\hat{\rho}_\text{his}^j) + \frac{1}{d'}\sum_{j=1}^{d'}KL(\rho||\hat{\rho}_\text{diff}^j).
\end{equation}
The final training objective combines the generation loss from the LLM and the SAE loss, including both reconstruction and sparsity terms:
\begin{equation}\small
    \mathcal{L}_\text{total} = \mathcal{L}_\text{gen} + \lambda \cdot (\mathcal{L}_\text{recon} + \gamma \cdot \mathcal{L}_\text{sparse}).
\end{equation}
where $\lambda$ and $\gamma$ balance the contributions of the SAE loss and the sparsity constraint, respectively.

\section{Experiment}

\begin{table*}[!ht]
    \centering
    \small
    \caption{Performance comparison between the baselines and our \ours across the three datasets. \textit{7B} and \textit{32B} represent the size of base LLMs. The best results are highlighted in \textbf{bold}, and the second-best results are \underline{underlined}. ``R-1'', ``MET.'', ``BL.'', and ``BS.'' respectively denote ROUGE-1, METEOR, BLEU, and BERTScore. Higher values indicate better performance across all metrics.}
    \renewcommand{\arraystretch}{1.2}
    \setlength{\tabcolsep}{3pt}
    \resizebox{0.99\textwidth}{!}{
    \begin{tabularx}{\textwidth}{>{\centering\arraybackslash}m{0.1cm} >{\centering\arraybackslash}m{1.88cm} *{12}{>{\centering\arraybackslash}m{0.93cm}}}
        \toprule
        \multicolumn{2}{c}{\textbf{Datasets ($\rightarrow$)}} & \multicolumn{4}{c}{\textbf{Books}} & \multicolumn{4}{c}{\textbf{Movies \& TV}} & \multicolumn{4}{c}{\textbf{CDs \& Vinyl}} \\
        \cmidrule(lr){3-6}
        \cmidrule(lr){7-10}
        \cmidrule(lr){11-14}
        \multicolumn{2}{c}{\textbf{Methods ($\downarrow$)}}  & R-1 & MET. & BL. & BS. & R-1 & MET. & BL. & BS. & R-1 & MET. & BL. & BS. \\
        \midrule
        \multirow{4}{*}{\textit{32B}} 
        & Non-Perso & 0.3025 & 0.1949 & 2.6728 & 0.4970 & 0.2608 & 0.1666 & 1.1226 & 0.4702 & 0.2765 & 0.1767 & 1.6597 & 0.4742 \\
        & RAG & \underline{0.3404} & 0.2735 & 6.8178 & \underline{0.5159} & \underline{0.2983} & 0.2142 & 2.8680 & 0.4822 & 0.3092 & 0.2177 & 3.1588 & 0.4868 \\
        & PAG & 0.3276 & 0.2830 & 6.8920 & 0.5051 & 0.2816 & 0.2130 & 2.7751 & 0.4746 & 0.2971 & 0.2215 & 3.2164 & 0.4787 \\
        & DPL & 0.3392 & \underline{0.3003} & \underline{7.7423} & 0.5156 & 0.2967 & \underline{0.2238} & \underline{3.2965} & \underline{0.4855} & \underline{0.3119} & \underline{0.2337} & \underline{3.8271} & \underline{0.4910} \\
        \midrule
        \multirow{6}{*}{\textit{7B}} 
        & Non-Perso & 0.2907 & 0.1735 & 1.9766 & 0.5004 & 0.2469 & 0.1503 & 0.7242 & 0.4713 & 0.2604 & 0.1561 & 1.0997 & 0.4753 \\
        & RAG & 0.3149 & 0.2101 & 3.6874 & 0.5083 & 0.2693 & 0.1701 & 1.3021 & 0.4787 & 0.2796 & 0.1733 & 1.6129 & 0.4824 \\
        & PAG & 0.3136 & 0.2378 & 4.6762 & 0.4992 & 0.2761 & 0.1905 & 1.9360 & 0.4735 & 0.2882 & 0.1979 & 2.4740 & 0.4789 \\
        & DPL & 0.3194 & 0.2459 & 5.6623 & 0.5050 & 0.2845 & 0.1958 & 2.2451 & 0.4795 & 0.2952 & 0.2003 & 2.6943 & 0.4838 \\
        & PPlug & 0.3033 & 0.2234 & 7.0469 & 0.5152 & 0.2530 & 0.1724 & 3.2291 & 0.4767 & 0.2619 & 0.1711 & 3.0753 & 0.4806 \\
        & \textbf{\ours (ours)} & \textbf{0.3745} & \textbf{0.3156} & \textbf{13.5300} & \textbf{0.5557} & \textbf{0.3092} & \textbf{0.2381} & \textbf{6.6835} & \textbf{0.5114} & \textbf{0.3165} & \textbf{0.2364} & \textbf{6.5166} & \textbf{0.5151} \\
        \bottomrule
    \end{tabularx}}
    \label{main_table}
\end{table*}

We conduct experiments in real-world datasets to answer the following research questions:

\begin{itemize}[leftmargin=*, topsep=2pt, itemsep=0pt]
    \item \textbf{RQ1}: How does \ours compare with baseline methods on the personalized text generation task?
    \item \textbf{RQ2}: What is the contribution of each individual component of \ours to its overall effectiveness?
    \item \textbf{RQ3}: What is the impact of the number of retrieved histories on the performance of \ours?
    \item \textbf{RQ4}: How does \ours perform under different levels of user uniqueness compared to DPL?
\end{itemize}

\subsection{Experimental Setup}

\noindent
\textbf{Datasets.} 
Building upon prior work, we focus on the representative task of item review generation for LLM personalization~\cite{amazon2018,reviewllm,longlamp,pgraphrag}.
Specifically, we adopt the Amazon Reviews 2023 dataset\footnote{\url{https://amazon-reviews-2023.github.io/}}~\cite{amazon2023} preprocessed by DPL\footnote{\url{https://huggingface.co/datasets/SnowCharmQ/DPL-main} \& \url{https://huggingface.co/datasets/SnowCharmQ/DPL-meta}}~\cite{dpl}, which covers three categories: \textit{Books}, \textit{Movies \& TV}, and \textit{CDs \& Vinyl}.
To maximize data utilization, we follow the setting of REST-PG~\cite{restpg} to train a unified model across categories.
For training, we retain each user's most recent interaction per category.
For validation, we randomly select 512 instances from the merged validation set across all three categories, while for testing, we follow the original test splits provided by DPL.
More details about the dataset are provided in Appendix~\ref{apd_dataset}.

\vspace{0.5em}
\noindent
\textbf{Baselines.}
We compare our proposed \ours with the following baseline methods. 
Further implementation details of all baselines can be found in Appendix~\ref{apd_baseline}.
\begin{itemize}[leftmargin=*, topsep=4pt, itemsep=0pt]
    \item \textbf{Non-Perso}: A non-personalized baseline that generates reviews using only item information, along with the review's title and rating.
    \item \textbf{RAG}~\cite{lamp}: A retrieval-based method that incorporates the user's history records to provide contextual personalization.
    \item \textbf{PAG}~\cite{pag}: An extension of RAG that summarizes the user's history records into a compact profile and combines it with retrieved content for higher-level personalization.
    \item \textbf{DPL}~\cite{dpl}: A prompt-based method that enhances personalization by explicitly comparing a user’s recent behavior with representative peers and summarizing the differences into a profile integrated into the LLM input.
    \item \textbf{PPlug}~\cite{pplug}: A plug-and-play approach that encodes user history into a dense embedding, which is projected into the LLM's input space to guide generation.
\end{itemize}

\vspace{0.5em}
\noindent
\textbf{Evaluation Metrics.}
Following previous works on personalized text generation~\cite{lamp,longlamp,stylevector,pgraphrag,reviewllm}, we evaluate all methods using ROUGE-1~\cite{rouge}, METEOR~\cite{meteor}, BLEU\footnote{We use the standard \texttt{SacreBLEU}~\cite{sacrebleu} library to calculate the BLEU score: \url{https://github.com/mjpost/sacrebleu}.}~\cite{bleu}, and BERTScore\footnote{We adopt the \texttt{led-base-16384}~\cite{longformer} model to obtain embeddings.}~\cite{bertscore}.

\vspace{0.5em}
\noindent
\textbf{Implementation Details.}
We utilize the \texttt{Qwen2.5-Instruct}\footnote{\url{https://huggingface.co/Qwen}}~\cite{qwen25} series models (7B and 32B) as backbone LLMs for baseline methods and \ours.
To retrieve user histories, we adopt a recency-based strategy, selecting the most recent history for each user.
Additionally, we employ \texttt{bge-m3}\footnote{\url{https://huggingface.co/BAAI/bge-m3}}~\cite{bgem3} as the embedding model to map user reviews into vector representations.
We train \ours for 5 epochs and select the checkpoint with the highest METEOR score on the validation set for testing.
For more details, please refer to Appendix~\ref{apd_implementation}.

\subsection{Main Results (RQ1)}

We first evaluate the overall performance of all compared methods.
Table~\ref{main_table} presents the main experimental results across three datasets, from which we draw the following observations:

\begin{itemize}[leftmargin=*, topsep=4pt, itemsep=0pt]

    \item \textbf{Incorporating context information significantly improves the model's capability for personalized text generation.}
    Methods like RAG and PAG leverage retrieved user information for generation, significantly outperforming the Non-Perso baseline.
    DPL further improves upon these by explicitly modeling inter-user differences, achieving the relatively best performance among all ICL-based methods.
    This shows that capturing user differences yields better personalization than simple relevance or summarization.
    \item \textbf{Scaling up the model size leads to stronger performance across different personalization methods.}
    For methods where both 7B and 32B models are evaluated, we observe consistent improvements across three metrics.
    This trend highlights the capacity of larger models to capture more nuanced personalization patterns.
    \item \textbf{Using a single soft prompt for user history, PPlug lacks informative signals and overlooks inter-user differences.}
    Although PPlug outperforms the Non-Perso baseline by introducing lightweight user modeling through the soft prompt, its gains remain limited.
    This limitation motivates our design of a more effective soft prompt strategy.
    \item \textbf{\ours consistently outperforms all baselines across datasets and metrics.}
    Despite operating on a much smaller model scale, \ours not only significantly outperforms all 7B-based methods, but also surpasses all baselines under the 32B backbone. Notably, averaged across three datasets, \ours yields relative improvements of 5.05\% in ROUGE-1, 4.21\% in METEOR, 82.59\% in BLEU, and 6.01\% in BERTScore compared to the strongest baseline. This substantial performance gain is primarily attributed to the integration of implicit modeling of user history and inter-user differences, which provides more informative and discriminative signals for personalization.
\end{itemize}

\subsection{Ablation Studies (RQ2)}

To better understand the contribution of different components in our personalization framework, we conduct extensive ablation studies from two perspectives: user embedding configuration and representation refinement.

We report METEOR scores on all three datasets here, and leave results for the other two metrics in Appendix~\ref{complete_ablation}.

\subsubsection{User Embedding Configuration}

To assess the effectiveness of incorporating different types of user embeddings, we conduct a detailed study comparing various configurations of personalized signals.
Specifically, we consider two types of embeddings: (1) user-specific embeddings (\textit{his\_emb}), which represent the user's past interactions, and (2) difference-aware embeddings (\textit{diff\_emb}), which encode inter-user differences by contrasting the target user's review history with those of other users.
We examine these embedding configurations individually and in combination, under two settings: with retrieved review text (\textit{w/ text}) and without it (\textit{w/o text}).

Results in Table~\ref{ablation_meteor_1} show that both \textit{his\_emb} and \textit{diff\_emb} individually outperform the non-personalized baseline, demonstrating the effectiveness of modeling both user history and inter-user differences.
Combining the two leads to further improvements, suggesting that user-specific embedding and difference-aware embedding capture complementary aspects of personalization.
Additionally, incorporating retrieved texts (\textit{w/ text}) consistently enhances all configurations, highlighting the benefit of contextual grounding.

\setlength{\dashlinedash}{6pt}
\begin{table}[t]
    \centering
    \small
    \caption{Ablation study on different configurations of user embeddings. \textit{his\_emb} and \textit{diff\_emb} denote user history and difference-aware embeddings. \textit{w/o text} and \textit{w/ text} refer to the exclusion or inclusion of retrieved review texts.}
    \renewcommand{\arraystretch}{1.2}
    \setlength{\tabcolsep}{4pt}
    \resizebox{0.45\textwidth}{!}{
    \begin{tabularx}{0.48\textwidth}{>{\centering\arraybackslash}m{0.3cm} >{\centering\arraybackslash}m{2.6cm} *{3}{>{\centering\arraybackslash}m{1.2cm}}}
        \toprule
        \multicolumn{2}{c}{\textbf{Datasets ($\rightarrow$)}} & \multirow{2}{*}{\textbf{Books}} & \multirow{2}{*}{\shortstack{\textbf{Movies} \\ \textbf{\& TV}}} & \multirow{2}{*}{\shortstack{\textbf{CDs \&} \\ \textbf{Vinyl}}} \\
        \multicolumn{2}{c}{\textbf{Methods ($\downarrow$)}} &  &  &  \\
        \midrule
        \multicolumn{2}{c}{Non-Perso-7B} & 0.1735 & 0.1503 & 0.1561 \\
        \noalign{\vskip 2pt}
        \hdashline
        \noalign{\vskip 2pt}
        \multirow{3}{*}{\rotatebox{90}{\textit{w/o text}}} 
        & his\_emb & 0.1718 & 0.1625 & 0.1711 \\
        & diff\_emb & 0.1839 & 0.1546 & 0.1616 \\
        & his\_emb + diff\_emb & 0.2227 & 0.1871 & 0.1853 \\
        \noalign{\vskip 2pt}
        \hdashline
        \noalign{\vskip 2pt}
        \multirow{3}{*}{\rotatebox{90}{\textit{w/ text}}}
        & his\_emb & 0.3110 & 0.2332 & 0.2268 \\
        & diff\_emb & 0.2781 & 0.2128 & 0.2108 \\
        & his\_emb + diff\_emb (ours) & \textbf{0.3156} & \textbf{0.2381} & \textbf{0.2364} \\
        \bottomrule
    \end{tabularx}
    }
    \label{ablation_meteor_1}
\end{table}

\begin{table}[t]
    \centering
    \small
    \caption{Ablation study on representation refinement. \textit{w/o DR} uses raw embeddings, \textit{w/ AE} uses a standard autoencoder, and \textit{w/ SAE} is our implementation.}
    \renewcommand{\arraystretch}{1.2}
    \setlength{\tabcolsep}{4pt}
    \resizebox{0.45\textwidth}{!}{
    \begin{tabularx}{0.45\textwidth}{>{\centering\arraybackslash}m{2.5cm} *{3}{>{\centering\arraybackslash}m{1.2cm}}}
        \toprule
        \textbf{Datasets ($\rightarrow$)} & \multirow{2}{*}{\textbf{Books}} & \multirow{2}{*}{\shortstack{\textbf{Movies} \\ \textbf{\& TV}}} & \multirow{2}{*}{\shortstack{\textbf{CDs \&} \\ \textbf{Vinyl}}} \\
        \textbf{Methods ($\downarrow$)} &  &  &  \\
        \midrule
        \textit{w/o DR} & 0.3016 & 0.2325 & 0.2283 \\
        \textit{w/ AE} & 0.2994 & 0.2350 & 0.2355 \\
        \textit{w/ SAE} (ours) & \textbf{0.3156} & \textbf{0.2381} & \textbf{0.2364} \\
        \bottomrule
    \end{tabularx}
    }
    \label{ablation_meteor_2}
\end{table}

\subsubsection{Representation Refinement}

We further evaluate the impact of different strategies for refining user embeddings before soft prompt injection.
Specifically, we compare three variants:
(1) \textit{w/o DR}, where raw high-dimensional embeddings are directly projected without dimensionality reduction, (2) \textit{w/ AE}, which uses a standard autoencoder for compression without sparsity, and (3) \textit{w/ SAE},  which applies our sparse autoencoder to introduce the sparsity constraint.

Table~\ref{ablation_meteor_2} shows that removing dimensionality reduction (\textit{w/o DR}) generally results in weaker performance.
While the standard autoencoder (\textit{w/ AE}) brings partial improvements on \textit{Movies \& TV} and \textit{CDs \& Vinyl} datasets, it does not consistently outperform the raw embedding variant, suggesting that compression alone is insufficient.
In contrast, we introduce a sparse autoencoder (\textit{w/ SAE}), achieving the best results across all datasets, highlighting the effectiveness of sparsity constraint in enhancing representation quality for personalization.

\begin{figure}[t]
    \centering
    \includegraphics[width=0.88\linewidth]{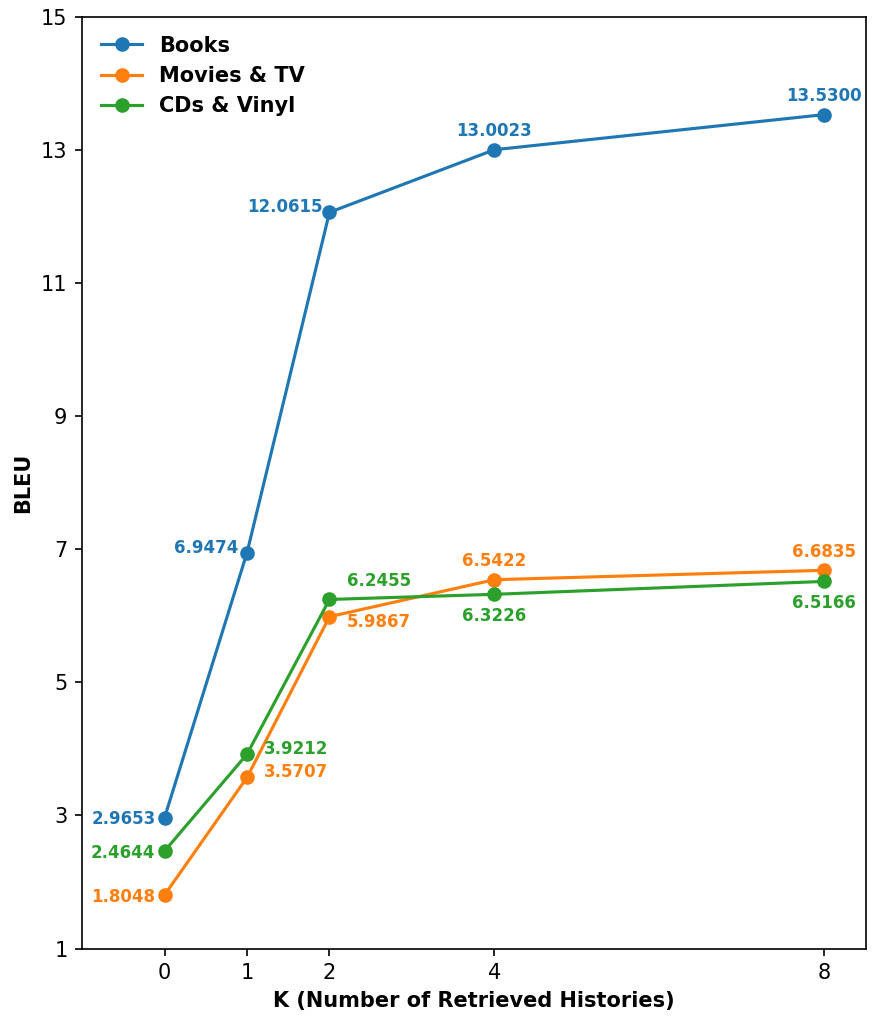}
    \caption{Effect of the number of retrieved user histories ($K$) on BLEU performance across datasets.}
    \label{num_retrieve}
\end{figure}

\begin{figure}[t]
    \centering
    \includegraphics[width=0.86\linewidth]{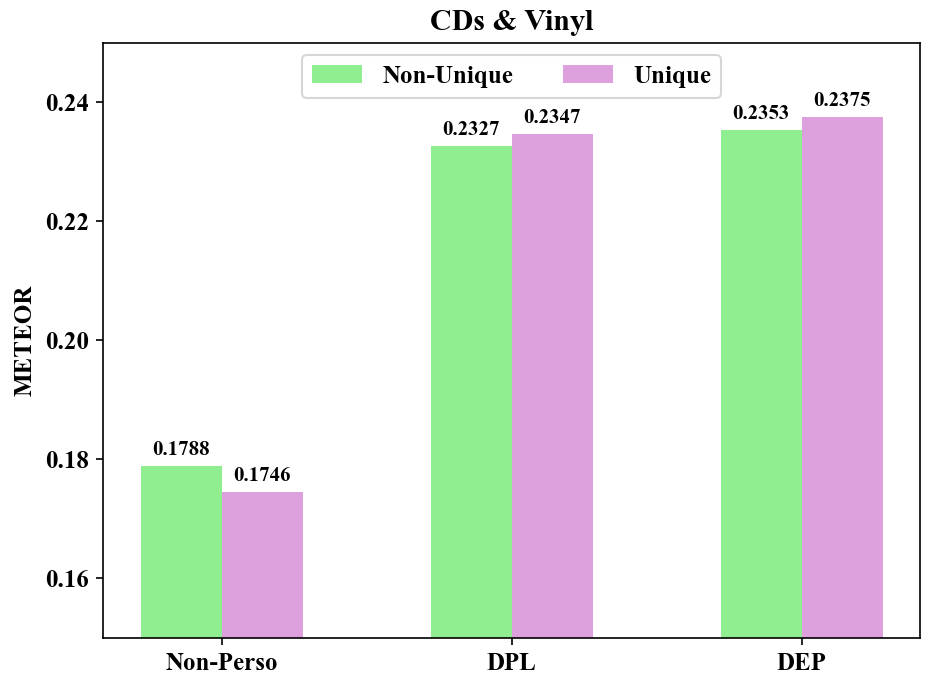}
    \caption{Results of the performance of \ours across different levels of uniqueness. The experiments are conducted on \textit{CDs \& Vinyl} and evaluated in METEOR.}
    \label{uniqueness}
    \vspace{-1em}
\end{figure}

\vspace{-3pt}
\subsection{In-Depth Analysis}

We conduct additional experiments to further study the design and effectiveness of our approach.

\subsubsection{Impact of History Number (RQ3)}

Figure~\ref{num_retrieve} shows how the number of retrieved user histories ($K$) affects the performance on BLEU across datasets.
A key observation is the substantial jump in performance from $K=0$ to $K=1$, which marks the transition from the non-personalized setting to the personalized framework of \ours.
This single-step increase highlights the substantial benefit of incorporating even one user-specific history with both the user-specific and difference-aware embeddings, demonstrating the effectiveness of our method once personalization is engaged.
As $K$ increases further, performance continues to improve, though with diminishing returns.

For a more comprehensive view, we provide more detailed results across other evaluation metrics and datasets in Appendix~\ref{history_number}.

\subsubsection{Impact of User Uniqueness (RQ4)}

Following the procedure in DPL, we further investigate how user uniqueness affects personalization performance. 
Similarly, we adopt a grouping strategy based on the user embedding derived from historical reviews.
Specifically, we compute the Euclidean distance between each user's review embedding and the global average embedding across all users, and divide users into two groups: the top 50\% as \textit{Unique} users and the bottom 50\% as \textit{Non-Unique} users.


As shown in Figure~\ref{uniqueness}, both DPL and \ours outperform the non-personalized baseline across user groups.
\ours consistently achieves the best results and maintains stable improvements for both \textit{Unique} and \textit{Non-Unique} users.
Similar to DPL, larger gains are observed in the \textit{Unique} group, highlighting the importance of modeling user distinctiveness.
Unlike DPL, which relies on prompt-level representations, \ours models inter-user differences in the latent space, enabling more compact and robust personalization, leading to better performance.

\section{Related Work}

Recent advancements~\cite{glm45,liangpearl,zhang2023autoalign,zhang-etal-2024-text,zhao2024pacar,lmtp,abstractthought,yao2025reasoning}  in large language models (LLMs) have demonstrated their strong generalization capabilities across diverse tasks~\cite{llmasajudge,collm,latentr3,dependeval,liu2025discrec,sarft,llm2rec,lastingbench,alphaedit,alphasteer}.
However, their ability to reflect personalized user intent remains limited.
Consequently, the personalization of LLMs has become a critical research direction, aiming to adapt general-purpose models to individual user preferences~\cite{prism,pad,igd,hipuid,prefeval,rlpa,nextquill,pmg,pigeon,drc}.
Among various approaches, the memory-retrieval framework~\cite{lamp} is widely adopted for its interpretability and scalability. 
It retrieves user-specific signals from interaction history to guide the model without changing its parameters. 
Methods under this framework generally fall into two types: retrieval-augmented generation (RAG) and profile-augmented generation (PAG).
RAG-based approaches retrieve relevant past interactions to construct a personalized prompt.
For example, HYDRA~\cite{hydra} employs a personalized reranker to refine retrieval quality, while PERAL~\cite{pearl} trains a retriever with a scale-calibrated objective to select useful information.
In contrast, PAG-based methods summarize the user's behavior into a condensed profile, which is then integrated into the prompt to guide generation~\cite{pag}.

Beyond retrieving individual histories, recent studies have explored incorporating other users' information as auxiliary signals to enhance individual personalization. 
CFRAG~\cite{cfrag}, Persona-DB~\cite{personadb}, and AP-Bots~\cite{apbots} borrow the concept of collaborative filtering~\cite{ncf,ngcf} to retrieve similar users' histories and incorporate them into the prompt to guide the generation.
DPL~\cite{dpl} further highlights that individual uniqueness lies in the differences from others and proposes to model such differences by formulating inter-user comparison as a language modeling task performed directly by the LLM.
While this method has shown promising results, modeling inter-user differences through prompt engineering poses challenges.
In contrast, our method shifts this process to the latent embedding space~\cite{uem,pplug,commer,userllm}, which avoids prompt-length constraints and enables more 
structured and nuanced modeling of user differences.

\section{Conclusion}

In this work, we propose \ours, a novel personalization framework that models inter-user differences in the latent embedding space to guide LLMs for personalized text generation.
Unlike prior approaches that rely only on prompt-level construction to integrate user histories and inter-user contrastive signals, our method jointly encodes both user-specific and difference-aware embeddings, and refines them through a sparse autoencoder to retain only task-relevant personalization cues.
These embeddings are then injected into a frozen LLM via soft prompts, enabling efficient personalization.
Experimental results across multiple domains show that \ours achieves state-of-the-art performance, especially for users with distinctive behavior patterns, confirming the effectiveness of latent inter-user difference modeling.
For future work, we plan to explore privacy-preserving inter-user comparison, real-time embedding updates, and extensions to tasks such as conversational agents.
\section*{Limitations}

While our proposed method \ours demonstrates strong performance in personalized text generation, it also introduces several limitations.
First, the method relies on sufficient user history to construct meaningful embeddings; in cold-start or data-sparse settings, its effectiveness may degrade.
Second, although more efficient than language-based comparison methods, the computation of difference-aware embeddings and the sparse autoencoder introduces additional overhead compared to standard prompting pipelines.
Lastly, our evaluation is centered on review generation, where preferences are explicit; adapting the approach to broader tasks like dialogue or recommendation requires further study.
\section*{Ethical Statements}

This work explores user-level personalization through the use of retrieved historical data and inter-user relational modeling.
While effective for improving generation quality, such approaches raise important ethical considerations.
In particular, accessing and processing users' historical interactions requires careful attention to data privacy, consent, and security.
Moreover, modeling inter-user differences may inadvertently expose sensitive behavioral patterns or amplify existing biases.

To mitigate these concerns, any real-world deployment of our method should incorporate privacy-preserving techniques such as anonymization, encryption, and transparent consent protocols.
Special care should be taken to avoid unintended inferences or misuse of user-level representations.

All experiments are conducted on publicly available datasets that have been preprocessed and released by prior work.
The original raw data is open-source and distributed under the MIT license.
We ensure that our use of the data adheres to established ethical standards and respects the original data usage guidelines.

We use AI assistants (e.g., ChatGPT) as auxiliary tools for writing refinement and coding support, while all research ideas, experimental designs, and final decisions are made by the authors.

\bibliography{custom}

\begin{thebibliography}{88}
\providecommand{\natexlab}[1]{#1}

\bibitem[{Achiam et~al.(2023)Achiam, Adler, Agarwal, Ahmad, Akkaya, Aleman, Almeida, Altenschmidt, Altman, Anadkat et~al.}]{gpt4}
Josh Achiam, Steven Adler, Sandhini Agarwal, Lama Ahmad, Ilge Akkaya, Florencia~Leoni Aleman, Diogo Almeida, Janko Altenschmidt, Sam Altman, Shyamal Anadkat, and 1 others. 2023.
\newblock Gpt-4 technical report.
\newblock \emph{arXiv preprint arXiv:2303.08774}.

\bibitem[{Au et~al.(2025)Au, Dimacali, Pedirappagari, Park, Dernoncourt, Wang, Kanakaris, Deilamsalehy, Rossi, and Ahmed}]{pgraphrag}
Steven Au, Cameron~J Dimacali, Ojasmitha Pedirappagari, Namyong Park, Franck Dernoncourt, Yu~Wang, Nikos Kanakaris, Hanieh Deilamsalehy, Ryan~A Rossi, and Nesreen~K Ahmed. 2025.
\newblock Personalized graph-based retrieval for large language models.
\newblock \emph{arXiv preprint arXiv:2501.02157}.

\bibitem[{Banerjee and Lavie(2005)}]{meteor}
Satanjeev Banerjee and Alon Lavie. 2005.
\newblock Meteor: An automatic metric for mt evaluation with improved correlation with human judgments.
\newblock In \emph{Proceedings of the acl workshop on intrinsic and extrinsic evaluation measures for machine translation and/or summarization}, pages 65--72.

\bibitem[{Beltagy et~al.(2020)Beltagy, Peters, and Cohan}]{longformer}
Iz~Beltagy, Matthew~E Peters, and Arman Cohan. 2020.
\newblock Longformer: The long-document transformer.
\newblock \emph{arXiv preprint arXiv:2004.05150}.

\bibitem[{Chen et~al.(2024{\natexlab{a}})Chen, Xiao, Zhang, Luo, Lian, and Liu}]{bgem3}
Jianlv Chen, Shitao Xiao, Peitian Zhang, Kun Luo, Defu Lian, and Zheng Liu. 2024{\natexlab{a}}.
\newblock Bge m3-embedding: Multi-lingual, multi-functionality, multi-granularity text embeddings through self-knowledge distillation.
\newblock \emph{arXiv preprint arXiv:2402.03216}.

\bibitem[{Chen et~al.(2024{\natexlab{b}})Chen, Liu, Huang, Wu, Liu, Jiang, Pu, Lei, Chen, Wang et~al.}]{personalizationsurvey1}
Jin Chen, Zheng Liu, Xu~Huang, Chenwang Wu, Qi~Liu, Gangwei Jiang, Yuanhao Pu, Yuxuan Lei, Xiaolong Chen, Xingmei Wang, and 1 others. 2024{\natexlab{b}}.
\newblock When large language models meet personalization: Perspectives of challenges and opportunities.
\newblock \emph{World Wide Web}, 27(4):42.

\bibitem[{Chen et~al.(2025{\natexlab{a}})Chen, Zhang, Luo, Chai, and Liu}]{pad}
Ruizhe Chen, Xiaotian Zhang, Meng Luo, Wenhao Chai, and Zuozhu Liu. 2025{\natexlab{a}}.
\newblock {PAD:} personalized alignment of llms at decoding-time.
\newblock In \emph{The Thirteenth International Conference on Learning Representations, {ICLR} 2025, 2025}.

\bibitem[{Chen et~al.(2025{\natexlab{b}})Chen, Zhao, Zhang, Zhang, Kawaguchi, Joty, Li, Chua, Shieh, and Zhang}]{abstractthought}
Yuxin Chen, Yiran Zhao, Yang Zhang, An~Zhang, Kenji Kawaguchi, Shafiq Joty, Junnan Li, Tat-Seng Chua, Michael~Qizhe Shieh, and Wenxuan Zhang. 2025{\natexlab{b}}.
\newblock The emergence of abstract thought in large language models beyond any language.
\newblock \emph{arXiv preprint arXiv:2506.09890}.

\bibitem[{Dao(2023)}]{flashattention2}
Tri Dao. 2023.
\newblock Flashattention-2: Faster attention with better parallelism and work partitioning.
\newblock \emph{arXiv preprint arXiv:2307.08691}.

\bibitem[{Doddapaneni et~al.(2024)Doddapaneni, Sayana, Jash, Sodhi, and Kuzmin}]{uem}
Sumanth Doddapaneni, Krishna Sayana, Ambarish Jash, Sukhdeep Sodhi, and Dima Kuzmin. 2024.
\newblock User embedding model for personalized language prompting.
\newblock In \emph{Proceedings of the 1st Workshop on Personalization of Generative AI Systems (PERSONALIZE 2024)}, pages 124--131.

\bibitem[{Du et~al.(2025)Du, Liu, Guo, Wang, Huang, Ni, and Li}]{dependeval}
Junjia Du, Yadi Liu, Hongcheng Guo, Jiawei Wang, Haojian Huang, Yunyi Ni, and Zhoujun Li. 2025.
\newblock {D}epend{E}val: Benchmarking {LLM}s for repository dependency understanding.
\newblock In \emph{Findings of the Association for Computational Linguistics: ACL 2025}, pages 7150--7179. Association for Computational Linguistics.

\bibitem[{Fang et~al.(2025{\natexlab{a}})Fang, Jiang, Wang, Ma, Shi, Wang, He, and Chua}]{alphaedit}
Junfeng Fang, Houcheng Jiang, Kun Wang, Yunshan Ma, Jie Shi, Xiang Wang, Xiangnan He, and Tat{-}Seng Chua. 2025{\natexlab{a}}.
\newblock Alphaedit: Null-space constrained knowledge editing for language models.
\newblock In \emph{The Thirteenth International Conference on Learning Representations, {ICLR} 2025, 2025}.

\bibitem[{Fang et~al.(2025{\natexlab{b}})Fang, Sun, Shi, and Gu}]{attentionrag}
Yixiong Fang, Tianran Sun, Yuling Shi, and Xiaodong Gu. 2025{\natexlab{b}}.
\newblock Attentionrag: Attention-guided context pruning in retrieval-augmented generation.
\newblock \emph{arXiv preprint arXiv:2503.10720}.

\bibitem[{Fang et~al.(2025{\natexlab{c}})Fang, Sun, Shi, Wang, and Gu}]{lastingbench}
Yixiong Fang, Tianran Sun, Yuling Shi, Min Wang, and Xiaodong Gu. 2025{\natexlab{c}}.
\newblock Lastingbench: Defend benchmarks against knowledge leakage.
\newblock \emph{arXiv preprint arXiv:2506.21614}.

\bibitem[{Glorot and Bengio(2010)}]{xavier}
Xavier Glorot and Yoshua Bengio. 2010.
\newblock Understanding the difficulty of training deep feedforward neural networks.
\newblock In \emph{Proceedings of the thirteenth international conference on artificial intelligence and statistics}, pages 249--256.

\bibitem[{Grattafiori et~al.(2024)Grattafiori, Dubey, Jauhri, Pandey, Kadian, Al-Dahle, Letman, Mathur, Schelten, Vaughan et~al.}]{llama3}
Aaron Grattafiori, Abhimanyu Dubey, Abhinav Jauhri, Abhinav Pandey, Abhishek Kadian, Ahmad Al-Dahle, Aiesha Letman, Akhil Mathur, Alan Schelten, Alex Vaughan, and 1 others. 2024.
\newblock The llama 3 herd of models.
\newblock \emph{arXiv preprint arXiv:2407.21783}.

\bibitem[{Guo et~al.(2025)Guo, Yang, Zhang, Song, Zhang, Xu, Zhu, Ma, Wang, Bi et~al.}]{deepseekr1}
Daya Guo, Dejian Yang, Haowei Zhang, Junxiao Song, Ruoyu Zhang, Runxin Xu, Qihao Zhu, Shirong Ma, Peiyi Wang, Xiao Bi, and 1 others. 2025.
\newblock Deepseek-r1: Incentivizing reasoning capability in llms via reinforcement learning.
\newblock \emph{arXiv preprint arXiv:2501.12948}.

\bibitem[{He et~al.(2017)He, Liao, Zhang, Nie, Hu, and Chua}]{ncf}
Xiangnan He, Lizi Liao, Hanwang Zhang, Liqiang Nie, Xia Hu, and Tat-Seng Chua. 2017.
\newblock Neural collaborative filtering.
\newblock In \emph{Proceedings of the 26th international conference on world wide web}, pages 173--182.

\bibitem[{He et~al.(2025)He, Liu, Zhang, Ma, and Chua}]{llm2rec}
Yingzhi He, Xiaohao Liu, An~Zhang, Yunshan Ma, and Tat-Seng Chua. 2025.
\newblock Llm2rec: Large language models are powerful embedding models for sequential recommendation.
\newblock In \emph{Proceedings of the 31st ACM SIGKDD Conference on Knowledge Discovery and Data Mining V. 2}, pages 896--907.

\bibitem[{Hou et~al.(2024)Hou, Li, He, Yan, Chen, and McAuley}]{amazon2023}
Yupeng Hou, Jiacheng Li, Zhankui He, An~Yan, Xiusi Chen, and Julian McAuley. 2024.
\newblock Bridging language and items for retrieval and recommendation.
\newblock \emph{arXiv preprint arXiv:2403.03952}.

\bibitem[{Huben et~al.(2024)Huben, Cunningham, Riggs, Ewart, and Sharkey}]{sae}
Robert Huben, Hoagy Cunningham, Logan Riggs, Aidan Ewart, and Lee Sharkey. 2024.
\newblock Sparse autoencoders find highly interpretable features in language models.
\newblock In \emph{The Twelfth International Conference on Learning Representations, {ICLR} 2024, 2024}.

\bibitem[{Irmak et~al.(2010)Irmak, Vallen, and Sen}]{diff3}
Caglar Irmak, Beth Vallen, and Sankar Sen. 2010.
\newblock You like what i like, but i don’t like what you like: Uniqueness motivations in product preferences.
\newblock \emph{Journal of Consumer Research}, 37(3):443--455.

\bibitem[{Izacard et~al.(2022)Izacard, Caron, Hosseini, Riedel, Bojanowski, Joulin, and Grave}]{contriever}
Gautier Izacard, Mathilde Caron, Lucas Hosseini, Sebastian Riedel, Piotr Bojanowski, Armand Joulin, and Edouard Grave. 2022.
\newblock Unsupervised dense information retrieval with contrastive learning.
\newblock \emph{Trans. Mach. Learn. Res.}, 2022.

\bibitem[{Kirk et~al.(2024)Kirk, Whitefield, R{\"{o}}ttger, Bean, Margatina, G{\'{o}}mez, Ciro, Bartolo, Williams, He, Vidgen, and Hale}]{prism}
Hannah~Rose Kirk, Alexander Whitefield, Paul R{\"{o}}ttger, Andrew~M. Bean, Katerina Margatina, Rafael~Mosquera G{\'{o}}mez, Juan Ciro, Max Bartolo, Adina Williams, He~He, Bertie Vidgen, and Scott Hale. 2024.
\newblock The {PRISM} alignment dataset: What participatory, representative and individualised human feedback reveals about the subjective and multicultural alignment of large language models.
\newblock In \emph{Advances in Neural Information Processing Systems 38: Annual Conference on Neural Information Processing Systems 2024, NeurIPS 2024, 2024}.

\bibitem[{Kumar et~al.(2024)Kumar, Viswanathan, Yerra, Salemi, Rossi, Dernoncourt, Deilamsalehy, Chen, Zhang, Agarwal et~al.}]{longlamp}
Ishita Kumar, Snigdha Viswanathan, Sushrita Yerra, Alireza Salemi, Ryan~A Rossi, Franck Dernoncourt, Hanieh Deilamsalehy, Xiang Chen, Ruiyi Zhang, Shubham Agarwal, and 1 others. 2024.
\newblock Longlamp: A benchmark for personalized long-form text generation.
\newblock \emph{arXiv preprint arXiv:2407.11016}.

\bibitem[{Kwon et~al.(2023)Kwon, Li, Zhuang, Sheng, Zheng, Yu, Gonzalez, Zhang, and Stoica}]{vllm}
Woosuk Kwon, Zhuohan Li, Siyuan Zhuang, Ying Sheng, Lianmin Zheng, Cody~Hao Yu, Joseph Gonzalez, Hao Zhang, and Ion Stoica. 2023.
\newblock Efficient memory management for large language model serving with pagedattention.
\newblock In \emph{Proceedings of the 29th Symposium on Operating Systems Principles}, pages 611--626.

\bibitem[{Li et~al.(2023)Li, Zhang, Mei, Wang, Hombaiah, Liang, and Bendersky}]{teachllm}
Cheng Li, Mingyang Zhang, Qiaozhu Mei, Yaqing Wang, Spurthi~Amba Hombaiah, Yi~Liang, and Michael Bendersky. 2023.
\newblock Teach llms to personalize--an approach inspired by writing education.
\newblock \emph{arXiv preprint arXiv:2308.07968}.

\bibitem[{Liang et~al.()Liang, Shen, Deng, Zhao, Liang, and Wong}]{liangpearl}
CHEN Liang, Li~Shen, Yang Deng, Xiaoyan Zhao, Bin Liang, and Kam-Fai Wong.
\newblock Pearl: Towards permutation-resilient llms.
\newblock In \emph{The Thirteenth International Conference on Learning Representations}.

\bibitem[{Lin(2004)}]{rouge}
Chin-Yew Lin. 2004.
\newblock Rouge: A package for automatic evaluation of summaries.
\newblock In \emph{Text summarization branches out}, pages 74--81.

\bibitem[{Lin et~al.(2025)Lin, Zhang, Zhao, Zhu, Feng, and Chua}]{igd}
Zijie Lin, Yang Zhang, Xiaoyan Zhao, Fengbin Zhu, Fuli Feng, and Tat-Seng Chua. 2025.
\newblock Igd: Token decisiveness modeling via information gain in llms for personalized recommendation.
\newblock \emph{arXiv preprint arXiv:2506.13229}.

\bibitem[{Liu et~al.(2025{\natexlab{a}})Liu, Bai, Zhao, Zhang, Feng, and Rong}]{liu2025discrec}
Chang Liu, Yimeng Bai, Xiaoyan Zhao, Yang Zhang, Fuli Feng, and Wenge Rong. 2025{\natexlab{a}}.
\newblock Discrec: Disentangled semantic-collaborative modeling for generative recommendation.
\newblock \emph{arXiv preprint arXiv:2506.15576}.

\bibitem[{Liu et~al.(2025{\natexlab{b}})Liu, Qiu, Li, Dai, Zhu, Hu, Yang, and King}]{personalizationsurvey4}
Jiahong Liu, Zexuan Qiu, Zhongyang Li, Quanyu Dai, Jieming Zhu, Minda Hu, Menglin Yang, and Irwin King. 2025{\natexlab{b}}.
\newblock A survey of personalized large language models: Progress and future directions.
\newblock \emph{arXiv preprint arXiv:2502.11528}.

\bibitem[{Liu et~al.(2025{\natexlab{c}})Liu, Zhu, Wang, Wei, Min, Lu, Wang, Yin, and Dou}]{pplug}
Jiongnan Liu, Yutao Zhu, Shuting Wang, Xiaochi Wei, Erxue Min, Yu~Lu, Shuaiqiang Wang, Dawei Yin, and Zhicheng Dou. 2025{\natexlab{c}}.
\newblock {LLM}s + persona-plug = personalized {LLM}s.
\newblock In \emph{Proceedings of the 63rd Annual Meeting of the Association for Computational Linguistics (Volume 1: Long Papers)}, pages 9373--9385. Association for Computational Linguistics.

\bibitem[{Liu et~al.(2025{\natexlab{d}})Liu, Xia, Zhao, Zhang, Yu, Su, Yang, Ng, and Chua}]{lmtp}
Xiaohao Liu, Xiaobo Xia, Weixiang Zhao, Manyi Zhang, Xianzhi Yu, Xiu Su, Shuo Yang, See-Kiong Ng, and Tat-Seng Chua. 2025{\natexlab{d}}.
\newblock L-mtp: Leap multi-token prediction beyond adjacent context for large language models.
\newblock \emph{arXiv preprint arXiv:2505.17505}.

\bibitem[{Loshchilov and Hutter(2019)}]{adamw}
Ilya Loshchilov and Frank Hutter. 2019.
\newblock Decoupled weight decay regularization.
\newblock In \emph{7th International Conference on Learning Representations, {ICLR} 2019, 2019}.

\bibitem[{Mok et~al.(2025)Mok, Kim, Park, and Yoon}]{hipuid}
Jisoo Mok, Ik-hwan Kim, Sangkwon Park, and Sungroh Yoon. 2025.
\newblock Exploring the potential of {LLM}s as personalized assistants: Dataset, evaluation, and analysis.
\newblock In \emph{Proceedings of the 63rd Annual Meeting of the Association for Computational Linguistics (Volume 1: Long Papers)}, pages 10212--10239. Association for Computational Linguistics.

\bibitem[{Mysore et~al.(2024)Mysore, Lu, Wan, Yang, Sarrafzadeh, Menezes, Baghaee, Gonzalez, Neville, and Safavi}]{pearl}
Sheshera Mysore, Zhuoran Lu, Mengting Wan, Longqi Yang, Bahareh Sarrafzadeh, Steve Menezes, Tina Baghaee, Emmanuel~Barajas Gonzalez, Jennifer Neville, and Tara Safavi. 2024.
\newblock Pearl: Personalizing large language model writing assistants with generation-calibrated retrievers.
\newblock In \emph{Proceedings of the 1st Workshop on Customizable NLP: Progress and Challenges in Customizing NLP for a Domain, Application, Group, or Individual (CustomNLP4U)}, pages 198--219.

\bibitem[{Ni et~al.(2019)Ni, Li, and McAuley}]{amazon2018}
Jianmo Ni, Jiacheng Li, and Julian McAuley. 2019.
\newblock Justifying recommendations using distantly-labeled reviews and fine-grained aspects.
\newblock In \emph{Proceedings of the 2019 conference on empirical methods in natural language processing and the 9th international joint conference on natural language processing (EMNLP-IJCNLP)}, pages 188--197.

\bibitem[{Ning et~al.(2025)Ning, Liu, Wu, Wu, Berlowitz, Prakash, Green, O'Banion, and Xie}]{userllm}
Lin Ning, Luyang Liu, Jiaxing Wu, Neo Wu, Devora Berlowitz, Sushant Prakash, Bradley Green, Shawn O'Banion, and Jun Xie. 2025.
\newblock User-llm: Efficient {LLM} contextualization with user embeddings.
\newblock In \emph{Companion Proceedings of the {ACM} on Web Conference 2025, {WWW} 2025, Sydney, NSW, Australia}, pages 1219--1223.

\bibitem[{Papineni et~al.(2002)Papineni, Roukos, Ward, and Zhu}]{bleu}
Kishore Papineni, Salim Roukos, Todd Ward, and Wei-Jing Zhu. 2002.
\newblock Bleu: a method for automatic evaluation of machine translation.
\newblock In \emph{Proceedings of the 40th annual meeting of the Association for Computational Linguistics}, pages 311--318.

\bibitem[{Paszke et~al.(2019)Paszke, Gross, Massa, Lerer, Bradbury, Chanan, Killeen, Lin, Gimelshein, Antiga et~al.}]{pytorch}
Adam Paszke, Sam Gross, Francisco Massa, Adam Lerer, James Bradbury, Gregory Chanan, Trevor Killeen, Zeming Lin, Natalia Gimelshein, Luca Antiga, and 1 others. 2019.
\newblock Pytorch: An imperative style, high-performance deep learning library.
\newblock \emph{Advances in neural information processing systems}, 32.

\bibitem[{Peng et~al.(2024)Peng, Liu, Xu, Yang, Shao, and Wang}]{reviewllm}
Qiyao Peng, Hongtao Liu, Hongyan Xu, Qing Yang, Minglai Shao, and Wenjun Wang. 2024.
\newblock Llm: Harnessing large language models for personalized review generation.
\newblock \emph{arXiv preprint arXiv:2407.07487}.

\bibitem[{Post(2018)}]{sacrebleu}
Matt Post. 2018.
\newblock A call for clarity in reporting {BLEU} scores.
\newblock In \emph{Proceedings of the Third Conference on Machine Translation: Research Papers}, pages 186--191.

\bibitem[{Qiu et~al.(2025)Qiu, Zhao, Zhang, Bai, Wang, Cheng, Feng, and Chua}]{dpl}
Yilun Qiu, Xiaoyan Zhao, Yang Zhang, Yimeng Bai, Wenjie Wang, Hong Cheng, Fuli Feng, and Tat-Seng Chua. 2025.
\newblock Measuring what makes you unique: Difference-aware user modeling for enhancing {LLM} personalization.
\newblock In \emph{Findings of the Association for Computational Linguistics: ACL 2025}, pages 21258--21277. Association for Computational Linguistics.

\bibitem[{Rajbhandari et~al.(2020)Rajbhandari, Rasley, Ruwase, and He}]{zero}
Samyam Rajbhandari, Jeff Rasley, Olatunji Ruwase, and Yuxiong He. 2020.
\newblock Zero: Memory optimizations toward training trillion parameter models.
\newblock In \emph{SC20: International Conference for High Performance Computing, Networking, Storage and Analysis}, pages 1--16. IEEE.

\bibitem[{Rasley et~al.(2020)Rasley, Rajbhandari, Ruwase, and He}]{deepspeed}
Jeff Rasley, Samyam Rajbhandari, Olatunji Ruwase, and Yuxiong He. 2020.
\newblock Deepspeed: System optimizations enable training deep learning models with over 100 billion parameters.
\newblock In \emph{Proceedings of the 26th ACM SIGKDD international conference on knowledge discovery \& data mining}, pages 3505--3506.

\bibitem[{Richardson et~al.(2023{\natexlab{a}})Richardson, Zhang, Gillespie, Kar, Singh, Raeesy, Khan, and Sethy}]{summ}
Chris Richardson, Yao Zhang, Kellen Gillespie, Sudipta Kar, Arshdeep Singh, Zeynab Raeesy, Omar~Zia Khan, and Abhinav Sethy. 2023{\natexlab{a}}.
\newblock Integrating summarization and retrieval for enhanced personalization via large language models.
\newblock \emph{arXiv preprint arXiv:2310.20081}.

\bibitem[{Richardson et~al.(2023{\natexlab{b}})Richardson, Zhang, Gillespie, Kar, Singh, Raeesy, Khan, and Sethy}]{pag}
Chris Richardson, Yao Zhang, Kellen Gillespie, Sudipta Kar, Arshdeep Singh, Zeynab Raeesy, Omar~Zia Khan, and Abhinav Sethy. 2023{\natexlab{b}}.
\newblock Integrating summarization and retrieval for enhanced personalization via large language models.
\newblock \emph{arXiv preprint arXiv:2310.20081}.

\bibitem[{Robertson et~al.(2009)Robertson, Zaragoza et~al.}]{bm25}
Stephen Robertson, Hugo Zaragoza, and 1 others. 2009.
\newblock The probabilistic relevance framework: Bm25 and beyond.
\newblock \emph{Foundations and Trends{\textregistered} in Information Retrieval}, 3(4):333--389.

\bibitem[{Salemi et~al.(2025)Salemi, Li, Zhang, Mei, Kong, Chen, Li, Bendersky, and Zamani}]{restpg}
Alireza Salemi, Cheng Li, Mingyang Zhang, Qiaozhu Mei, Weize Kong, Tao Chen, Zhuowan Li, Michael Bendersky, and Hamed Zamani. 2025.
\newblock Reasoning-enhanced self-training for long-form personalized text generation.
\newblock \emph{arXiv preprint arXiv:2501.04167}.

\bibitem[{Salemi et~al.(2024)Salemi, Mysore, Bendersky, and Zamani}]{lamp}
Alireza Salemi, Sheshera Mysore, Michael Bendersky, and Hamed Zamani. 2024.
\newblock {L}a{MP}: When large language models meet personalization.
\newblock In \emph{Proceedings of the 62nd Annual Meeting of the Association for Computational Linguistics (Volume 1: Long Papers)}, pages 7370--7392.

\bibitem[{Shen et~al.(2024{\natexlab{a}})Shen, Chen, Shao, Guan, and Nie}]{mome}
Leyang Shen, Gongwei Chen, Rui Shao, Weili Guan, and Liqiang Nie. 2024{\natexlab{a}}.
\newblock Mome: Mixture of multimodal experts for generalist multimodal large language models.
\newblock In \emph{Advances in Neural Information Processing Systems 38: Annual Conference on Neural Information Processing Systems 2024, NeurIPS 2024, Vancouver, BC, Canada, December 10 - 15, 2024}.

\bibitem[{Shen et~al.(2024{\natexlab{b}})Shen, Zhang, Zhao, Zhu, and Xiao}]{pmg}
Xiaoteng Shen, Rui Zhang, Xiaoyan Zhao, Jieming Zhu, and Xi~Xiao. 2024{\natexlab{b}}.
\newblock Pmg: Personalized multimodal generation with large language models.
\newblock In \emph{Proceedings of the ACM Web Conference 2024}, pages 3833--3843.

\bibitem[{Shen et~al.(2024{\natexlab{c}})Shen, Song, Tan, Zhang, Ren, Yuan, Lu, Li, and Zhuang}]{taskbench}
Yongliang Shen, Kaitao Song, Xu~Tan, Wenqi Zhang, Kan Ren, Siyu Yuan, Weiming Lu, Dongsheng Li, and Yueting Zhuang. 2024{\natexlab{c}}.
\newblock Taskbench: Benchmarking large language models for task automation.
\newblock In \emph{Advances in Neural Information Processing Systems 38: Annual Conference on Neural Information Processing Systems 2024, NeurIPS 2024, 2024}.

\bibitem[{Sheng et~al.(2025)Sheng, Shen, Zhao, Fang, Liu, Liang, Wang, Zhang, and Chua}]{alphasteer}
Leheng Sheng, Changshuo Shen, Weixiang Zhao, Junfeng Fang, Xiaohao Liu, Zhenkai Liang, Xiang Wang, An~Zhang, and Tat-Seng Chua. 2025.
\newblock Alphasteer: Learning refusal steering with principled null-space constraint.
\newblock \emph{arXiv preprint arXiv:2506.07022}.

\bibitem[{Shi et~al.(2025)Shi, Xu, Zhang, Zang, Zheng, Song, and Li}]{cfrag}
Teng Shi, Jun Xu, Xiao Zhang, Xiaoxue Zang, Kai Zheng, Yang Song, and Han Li. 2025.
\newblock Retrieval augmented generation with collaborative filtering for personalized text generation.
\newblock \emph{arXiv preprint arXiv:2504.05731}.

\bibitem[{Snyder and Fromkin(1977)}]{diff1}
Charles~R Snyder and Howard~L Fromkin. 1977.
\newblock Abnormality as a positive characteristic: The development and validation of a scale measuring need for uniqueness.
\newblock \emph{Journal of Abnormal Psychology}, 86(5):518.

\bibitem[{Snyder and Fromkin(2012)}]{diff2}
Charles~R Snyder and Howard~L Fromkin. 2012.
\newblock \emph{Uniqueness: The human pursuit of difference}.
\newblock Springer Science \& Business Media.

\bibitem[{Sun et~al.(2025)Sun, Yang, Gangi~Reddy, Fung, Chan, Small, Zhai, and Ji}]{personadb}
Chenkai Sun, Ke~Yang, Revanth Gangi~Reddy, Yi~Fung, Hou~Pong Chan, Kevin Small, ChengXiang Zhai, and Heng Ji. 2025.
\newblock Persona-{DB}: Efficient large language model personalization for response prediction with collaborative data refinement.
\newblock In \emph{Proceedings of the 31st International Conference on Computational Linguistics}, pages 281--296.

\bibitem[{Team et~al.(2025)Team, Kamath, Ferret, Pathak, Vieillard, Merhej, Perrin, Matejovicova, Ram{\'e}, Rivi{\`e}re et~al.}]{gemma3}
Gemma Team, Aishwarya Kamath, Johan Ferret, Shreya Pathak, Nino Vieillard, Ramona Merhej, Sarah Perrin, Tatiana Matejovicova, Alexandre Ram{\'e}, Morgane Rivi{\`e}re, and 1 others. 2025.
\newblock Gemma 3 technical report.
\newblock \emph{arXiv preprint arXiv:2503.19786}.

\bibitem[{Wang et~al.(2019)Wang, He, Wang, Feng, and Chua}]{ngcf}
Xiang Wang, Xiangnan He, Meng Wang, Fuli Feng, and Tat-Seng Chua. 2019.
\newblock Neural graph collaborative filtering.
\newblock In \emph{Proceedings of the 42nd international ACM SIGIR conference on Research and development in Information Retrieval}, pages 165--174.

\bibitem[{Wolf et~al.(2020)Wolf, Debut, Sanh, Chaumond, Delangue, Moi, Cistac, Rault, Louf, Funtowicz et~al.}]{transformers}
Thomas Wolf, Lysandre Debut, Victor Sanh, Julien Chaumond, Clement Delangue, Anthony Moi, Pierric Cistac, Tim Rault, R{\'e}mi Louf, Morgan Funtowicz, and 1 others. 2020.
\newblock Transformers: State-of-the-art natural language processing.
\newblock In \emph{Proceedings of the 2020 conference on empirical methods in natural language processing: system demonstrations}, pages 38--45.

\bibitem[{Xu et~al.(2025{\natexlab{a}})Xu, Wang, Zhang, Tang, Yan, Feng, and He}]{pigeon}
Yiyan Xu, Wenjie Wang, Yang Zhang, Biao Tang, Peng Yan, Fuli Feng, and Xiangnan He. 2025{\natexlab{a}}.
\newblock Personalized image generation with large multimodal models.
\newblock In \emph{Proceedings of the ACM on Web Conference 2025}, pages 264--274.

\bibitem[{Xu et~al.(2025{\natexlab{b}})Xu, Zhang, Salemi, Hu, Wang, Feng, Zamani, He, and Chua}]{personalizationsurvey3}
Yiyan Xu, Jinghao Zhang, Alireza Salemi, Xinting Hu, Wenjie Wang, Fuli Feng, Hamed Zamani, Xiangnan He, and Tat-Seng Chua. 2025{\natexlab{b}}.
\newblock Personalized generation in large model era: A survey.
\newblock In \emph{Proceedings of the 63rd Annual Meeting of the Association for Computational Linguistics (Volume 1: Long Papers)}, pages 24607--24649.

\bibitem[{Xu et~al.(2025{\natexlab{c}})Xu, Zheng, Wang, Zhu, Hu, Zhang, Feng, and Chua}]{drc}
Yiyan Xu, Wuqiang Zheng, Wenjie Wang, Fengbin Zhu, Xinting Hu, Yang Zhang, Fuli Feng, and Tat-Seng Chua. 2025{\natexlab{c}}.
\newblock Drc: Enhancing personalized image generation via disentangled representation composition.
\newblock \emph{arXiv preprint arXiv:2504.17349}.

\bibitem[{Yang et~al.(2025)Yang, Li, Yang, Zhang, Hui, Zheng, Yu, Gao, Huang, Lv et~al.}]{qwen3}
An~Yang, Anfeng Li, Baosong Yang, Beichen Zhang, Binyuan Hui, Bo~Zheng, Bowen Yu, Chang Gao, Chengen Huang, Chenxu Lv, and 1 others. 2025.
\newblock Qwen3 technical report.
\newblock \emph{arXiv preprint arXiv:2505.09388}.

\bibitem[{Yang et~al.(2024)Yang, Yang, Zhang, Hui, Zheng, Yu, Li, Liu, Huang, Wei et~al.}]{qwen25}
An~Yang, Baosong Yang, Beichen Zhang, Binyuan Hui, Bo~Zheng, Bowen Yu, Chengyuan Li, Dayiheng Liu, Fei Huang, Haoran Wei, and 1 others. 2024.
\newblock Qwen2. 5 technical report.
\newblock \emph{arXiv preprint arXiv:2412.15115}.

\bibitem[{Yao et~al.(2023)Yao, Zhao, Yu, Du, Shafran, Narasimhan, and Cao}]{react}
Shunyu Yao, Jeffrey Zhao, Dian Yu, Nan Du, Izhak Shafran, Karthik~R. Narasimhan, and Yuan Cao. 2023.
\newblock React: Synergizing reasoning and acting in language models.
\newblock In \emph{The Eleventh International Conference on Learning Representations, {ICLR} 2023, 2023}.

\bibitem[{Yao et~al.(2025)Yao, Liu, Chen, Chen, Fang, Hou, Li, and Chua}]{yao2025reasoning}
Zijun Yao, Yantao Liu, Yanxu Chen, Jianhui Chen, Junfeng Fang, Lei Hou, Juanzi Li, and Tat-Seng Chua. 2025.
\newblock Are reasoning models more prone to hallucination?
\newblock \emph{arXiv preprint arXiv:2505.23646}.

\bibitem[{Yazan et~al.(2025)Yazan, Verberne, and Situmeang}]{apbots}
Mert Yazan, Suzan Verberne, and Frederik Situmeang. 2025.
\newblock Improving rag for personalization with author features and contrastive examples.
\newblock In \emph{European Conference on Information Retrieval}, pages 408--416. Springer.

\bibitem[{Zeldes et~al.(2025)Zeldes, Zait, Labzovsky, Karmon, and Farkash}]{commer}
Yoel Zeldes, Amir Zait, Ilia Labzovsky, Danny Karmon, and Efrat Farkash. 2025.
\newblock Commer: a framework for compressing and merging user data for personalization.
\newblock \emph{arXiv preprint arXiv:2501.03276}.

\bibitem[{Zeng et~al.(2025)Zeng, Lv, Zheng, Hou, Chen, Xie, Wang, Yin, Zeng, Zhang et~al.}]{glm45}
Aohan Zeng, Xin Lv, Qinkai Zheng, Zhenyu Hou, Bin Chen, Chengxing Xie, Cunxiang Wang, Da~Yin, Hao Zeng, Jiajie Zhang, and 1 others. 2025.
\newblock Glm-4.5: Agentic, reasoning, and coding (arc) foundation models.
\newblock \emph{arXiv preprint arXiv:2508.06471}.

\bibitem[{Zhang et~al.(2025{\natexlab{a}})Zhang, Liu, Wang, Liu, Wu, Wang, and Chua}]{stylevector}
Jinghao Zhang, Yuting Liu, Wenjie Wang, Qiang Liu, Shu Wu, Liang Wang, and Tat-Seng Chua. 2025{\natexlab{a}}.
\newblock Personalized text generation with contrastive activation steering.
\newblock In \emph{Proceedings of the 63rd Annual Meeting of the Association for Computational Linguistics (Volume 1: Long Papers)}, pages 7128--7141. Association for Computational Linguistics.

\bibitem[{Zhang et~al.(2023)Zhang, Su, Trisedya, Zhao, Yang, Cheng, and Qi}]{zhang2023autoalign}
Rui Zhang, Yixin Su, Bayu~Distiawan Trisedya, Xiaoyan Zhao, Min Yang, Hong Cheng, and Jianzhong Qi. 2023.
\newblock Autoalign: Fully automatic and effective knowledge graph alignment enabled by large language models.
\newblock \emph{IEEE Transactions on Knowledge and Data Engineering}, 36(6):2357--2371.

\bibitem[{Zhang et~al.(2020)Zhang, Kishore, Wu, Weinberger, and Artzi}]{bertscore}
Tianyi Zhang, Varsha Kishore, Felix Wu, Kilian~Q. Weinberger, and Yoav Artzi. 2020.
\newblock Bertscore: Evaluating text generation with {BERT}.
\newblock In \emph{8th International Conference on Learning Representations, {ICLR} 2020, 2020}.

\bibitem[{Zhang et~al.(2024{\natexlab{a}})Zhang, Bao, Yan, Wang, Feng, and He}]{zhang-etal-2024-text}
Yang Zhang, Keqin Bao, Ming Yan, Wenjie Wang, Fuli Feng, and Xiangnan He. 2024{\natexlab{a}}.
\newblock Text-like encoding of collaborative information in large language models for recommendation.
\newblock In \emph{Proceedings of the 62nd Annual Meeting of the Association for Computational Linguistics (Volume 1: Long Papers)}. Association for Computational Linguistics.

\bibitem[{Zhang et~al.(2025{\natexlab{b}})Zhang, Feng, Zhang, Bao, Wang, and He}]{collm}
Yang Zhang, Fuli Feng, Jizhi Zhang, Keqin Bao, Qifan Wang, and Xiangnan He. 2025{\natexlab{b}}.
\newblock Collm: Integrating collaborative embeddings into large language models for recommendation.
\newblock \emph{IEEE Transactions on Knowledge and Data Engineering}.

\bibitem[{Zhang et~al.(2025{\natexlab{c}})Zhang, Xu, Zhao, Wang, Feng, He, and Chua}]{latentr3}
Yang Zhang, Wenxin Xu, Xiaoyan Zhao, Wenjie Wang, Fuli Feng, Xiangnan He, and Tat-Seng Chua. 2025{\natexlab{c}}.
\newblock Reinforced latent reasoning for llm-based recommendation.
\newblock \emph{arXiv preprint arXiv:2505.19092}.

\bibitem[{Zhang et~al.(2024{\natexlab{b}})Zhang, Rossi, Kveton, Shao, Yang, Zamani, Dernoncourt, Barrow, Yu, Kim et~al.}]{personalizationsurvey2}
Zhehao Zhang, Ryan~A Rossi, Branislav Kveton, Yijia Shao, Diyi Yang, Hamed Zamani, Franck Dernoncourt, Joe Barrow, Tong Yu, Sungchul Kim, and 1 others. 2024{\natexlab{b}}.
\newblock Personalization of large language models: A survey.
\newblock \emph{arXiv preprint arXiv:2411.00027}.

\bibitem[{Zhao et~al.(2025{\natexlab{a}})Zhao, Hong, Liu, Hazarika, and Lin}]{prefeval}
Siyan Zhao, Mingyi Hong, Yang Liu, Devamanyu Hazarika, and Kaixiang Lin. 2025{\natexlab{a}}.
\newblock Do llms recognize your preferences? evaluating personalized preference following in llms.
\newblock In \emph{The Thirteenth International Conference on Learning Representations, {ICLR} 2025, 2025}.

\bibitem[{Zhao et~al.(2025{\natexlab{b}})Zhao, Hu, Deng, Guo, Sui, Han, Zhang, Zhao, Qin, Chua, and Liu}]{sarft}
Weixiang Zhao, Yulin Hu, Yang Deng, Jiahe Guo, Xingyu Sui, Xinyang Han, An~Zhang, Yanyan Zhao, Bing Qin, Tat-Seng Chua, and Ting Liu. 2025{\natexlab{b}}.
\newblock Beware of your po! measuring and mitigating {AI} safety risks in role-play fine-tuning of {LLM}s.
\newblock In \emph{Proceedings of the 63rd Annual Meeting of the Association for Computational Linguistics (Volume 1: Long Papers)}, pages 11112--11137. Association for Computational Linguistics.

\bibitem[{Zhao et~al.(2025{\natexlab{c}})Zhao, Sui, Hu, Guo, Liu, Li, Zhao, Qin, and Liu}]{rlpa}
Weixiang Zhao, Xingyu Sui, Yulin Hu, Jiahe Guo, Haixiao Liu, Biye Li, Yanyan Zhao, Bing Qin, and Ting Liu. 2025{\natexlab{c}}.
\newblock Teaching language models to evolve with users: Dynamic profile modeling for personalized alignment.
\newblock \emph{arXiv preprint arXiv:2505.15456}.

\bibitem[{Zhao et~al.(2025{\natexlab{d}})Zhao, Deng, Wang, Cheng, Zhang, Ng, Chua et~al.}]{zhao2025exploring}
Xiaoyan Zhao, Yang Deng, Wenjie Wang, Hong Cheng, Rui Zhang, See-Kiong Ng, Tat-Seng Chua, and 1 others. 2025{\natexlab{d}}.
\newblock Exploring the impact of personality traits on conversational recommender systems: A simulation with large language models.
\newblock \emph{arXiv preprint arXiv:2504.12313}.

\bibitem[{Zhao et~al.(2024{\natexlab{a}})Zhao, Deng, Yang, Wang, Zhang, Cheng, Lam, Shen, and Xu}]{zhao2024comprehensive}
Xiaoyan Zhao, Yang Deng, Min Yang, Lingzhi Wang, Rui Zhang, Hong Cheng, Wai Lam, Ying Shen, and Ruifeng Xu. 2024{\natexlab{a}}.
\newblock A comprehensive survey on relation extraction: Recent advances and new frontiers.
\newblock \emph{ACM Computing Surveys}, 56(11):1--39.

\bibitem[{Zhao et~al.(2024{\natexlab{b}})Zhao, Wang, Wang, Cheng, Zhang, and Wong}]{zhao2024pacar}
Xiaoyan Zhao, Lingzhi Wang, Zhanghao Wang, Hong Cheng, Rui Zhang, and Kam-Fai Wong. 2024{\natexlab{b}}.
\newblock Pacar: Automated fact-checking with planning and customized action reasoning using large language models.
\newblock In \emph{Proceedings of the 2024 Joint International Conference on Computational Linguistics, Language Resources and Evaluation (LREC-COLING 2024)}, pages 12564--12573.

\bibitem[{Zhao et~al.(2025{\natexlab{e}})Zhao, You, Zhang, Wang, Cheng, Feng, Ng, and Chua}]{nextquill}
Xiaoyan Zhao, Juntao You, Yang Zhang, Wenjie Wang, Hong Cheng, Fuli Feng, See-Kiong Ng, and Tat-Seng Chua. 2025{\natexlab{e}}.
\newblock Nextquill: Causal preference modeling for enhancing llm personalization.
\newblock \emph{arXiv preprint arXiv:2506.02368}.

\bibitem[{Zheng et~al.(2023)Zheng, Chiang, Sheng, Zhuang, Wu, Zhuang, Lin, Li, Li, Xing, Zhang, Gonzalez, and Stoica}]{llmasajudge}
Lianmin Zheng, Wei{-}Lin Chiang, Ying Sheng, Siyuan Zhuang, Zhanghao Wu, Yonghao Zhuang, Zi~Lin, Zhuohan Li, Dacheng Li, Eric~P. Xing, Hao Zhang, Joseph~E. Gonzalez, and Ion Stoica. 2023.
\newblock Judging llm-as-a-judge with mt-bench and chatbot arena.
\newblock In \emph{Advances in Neural Information Processing Systems 36: Annual Conference on Neural Information Processing Systems 2023, NeurIPS 2023, 2023}.

\bibitem[{Zhuang et~al.(2024)Zhuang, Sun, Yu, Qiang, Wang, Zhang, and Dai}]{hydra}
Yuchen Zhuang, Haotian Sun, Yue Yu, Rushi Qiang, Qifan Wang, Chao Zhang, and Bo~Dai. 2024.
\newblock Hydra: Model factorization framework for black-box llm personalization.
\newblock In \emph{Advances in Neural Information Processing Systems}, volume~37, pages 100783--100815.

\end{thebibliography}

\appendix

\section{Dataset Details}\label{apd_dataset}

In this paper, we focus on the task of review generation. 
Specifically, we adopt the Amazon~\cite{amazon2023} dataset preprocessed by DPL~\cite{dpl}.
We select each user's most recent interaction from the training sets of the three categories and merge them into a unified training dataset, which is used to train the model.
For validation, we also aggregate the three categories and randomly sample 512 instances.
For testing, we directly use the test splits preprocessed by DPL.
During data preprocessing, we construct complete prompts as model inputs by concatenating the target item title, target item description, output review title, output review rating, and the retrieved user's past reviews.
For clarity, we provide an example of the dataset preprocessed by DPL as shown in Figure~\ref{dataset_demo}, and dataset statistics after processing are summarized in Table~\ref{dataset_stat}.

\section{Baseline Details}\label{apd_baseline}

We compare our proposed \ours with several baseline methods. 
The comparison between different baselines and our method is shown in Table~\ref{method_comparison}.
In this section, we further introduce each baseline method in detail:

\begin{itemize}[leftmargin=*]
    \item \textbf{Non-Perso}: This method generates reviews without leveraging any user-specific information. The input to the model includes only the item's title and description, along with the output review's rating and title.
    \item \textbf{RAG}~\cite{lamp}: This method uses a simple recency-based retrieval strategy to select the most recent reviews from the user's history.
    The retrieved reviews are then directly formatted and incorporated into the LLM's input to provide contextual personalization.
\begin{figure}[t]
    \centering    
    \includegraphics[width=1\linewidth]{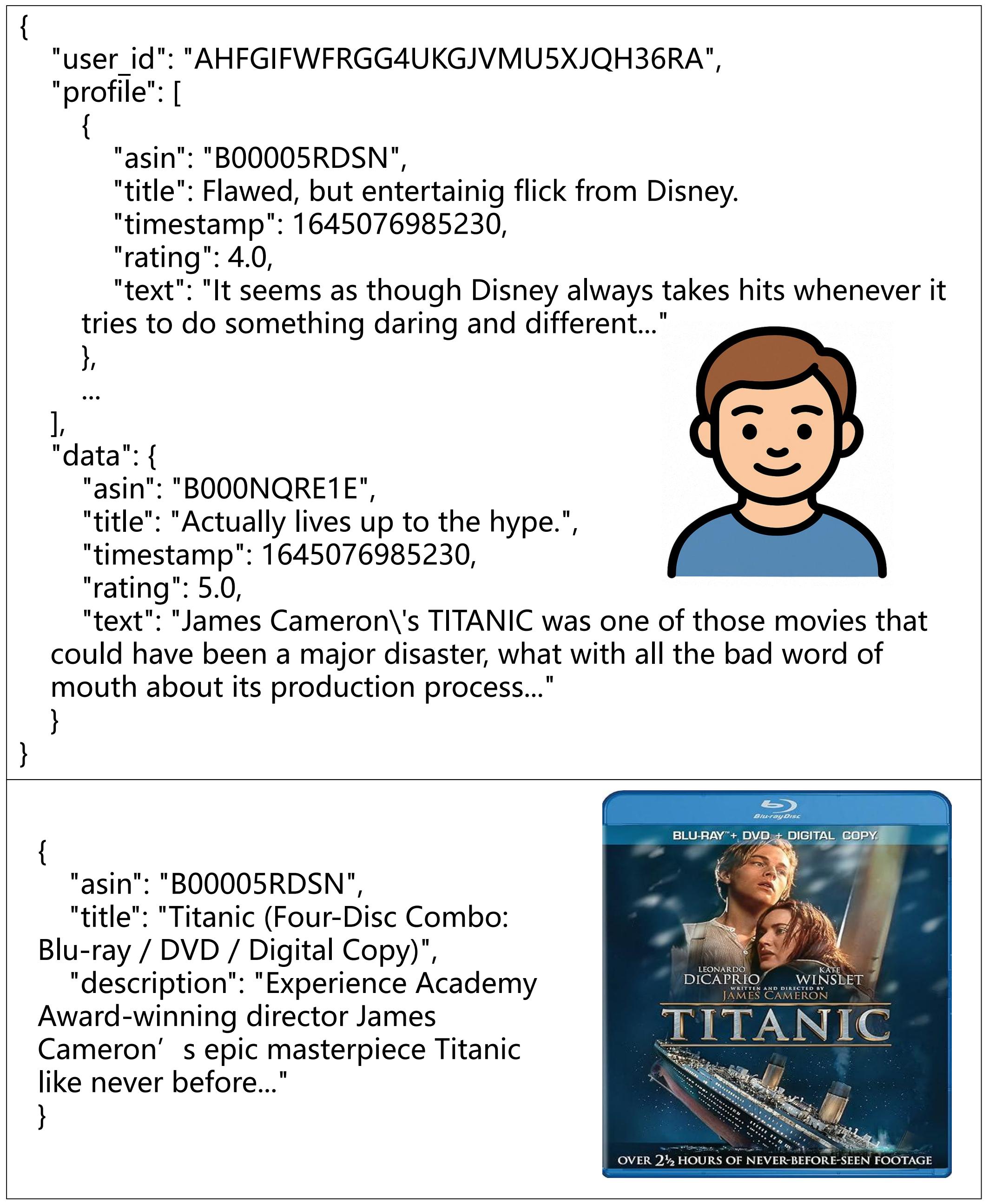}
    \caption{An example of the user review from the main dataset (above) and the corresponding item from the meta dataset (below).}
    \label{dataset_demo}
\end{figure}
\begin{table}[t]
    \centering
    \caption{Overview of dataset statistics across the three benchmark categories.}
    \renewcommand{\arraystretch}{1.1}
    \setlength{\tabcolsep}{6pt} 
    \resizebox{0.48\textwidth}{!}{
    \begin{tabular}{ccccc}
        \toprule
        \multicolumn{2}{c}{\textbf{Categories ($\downarrow$)}} & \#data & Profile Size & Output Length \\
        \midrule
        \multicolumn{2}{c}{\textit{Training Dataset}} & 3996 & 37.47$\pm$33.53 & 1608.82$\pm$1476.99 \\
        \midrule
        \multicolumn{2}{c}{\textit{Validation Dataset}} & 512 & 39.14$\pm$36.01 & 1557.29$\pm$1378.43 \\
        \midrule
        \multirow{3}{*}{\textit{\shortstack{Test\\Dataset}}} 
        & \textbf{Books} & 317 & 34.84$\pm$22.55 & 1194.90$\pm$802.44 \\
        & \textbf{Movies \& TV } & 1925 & 41.11$\pm$35.90 & 1704.61$\pm$1752.44 \\
        & \textbf{CDs \& Vinyl } & 1754 & 38.50$\pm$32.37 & 1600.04$\pm$1419.89 \\
        \bottomrule
    \end{tabular}}
    \label{dataset_stat}
\end{table}
\begin{table*}[t]
    \caption{We provide a comparison between the different baseline methods and our proposed \ours, focusing on the following aspects:
    (1) retrieval augmentation, (2) embedded representation, and (3) inter-user difference.}
    \label{method_comparison}
    \centering
    \renewcommand{\arraystretch}{1.08}
    \resizebox{0.96\textwidth}{!}{\begin{tabular}{>{\centering\arraybackslash}m{2.2cm} cc  cc  c}
        \toprule
        \textbf{Methods ($\downarrow$)}  & Retrieval Augmentation & Embedded Representation & Inter-User Difference \\
        \midrule
        Non-Perso  & \xmark    & \xmark         & \xmark            \\
        RAG         & \cmark    & \xmark         & \xmark           \\
        PAG         & \cmark    & \xmark         & \xmark         \\
        DPL         & \cmark    & \xmark         & \cmark         \\
        PPlug         & \xmark    & \cmark         & \xmark         \\
        \textbf{\ours}  & \cmark    & \cmark         & \cmark         \\
        \bottomrule
    \end{tabular}
    }
\end{table*}
    \item \textbf{PAG}~\cite{pag}: Building upon RAG, this method first summarizes the most recent reviews from the user's history into a compact profile.
    The generated profile, along with the retrieved records, is included in the input to the LLM, allowing it to generate personalized reviews guided by a higher-level understanding of the user.
    \item \textbf{DPL}~\cite{dpl}: The method prompts the LLM to find inter-user differences by comparing the target user's most recent interactions with representative users selected via clustering from predefined dimensions (\eg writing, emitional tone, and semantics), and summarizes them with the user's history to form a user profile.
    This profile, along with recent reviews, is incorporated into the model input to enhance generation.
    To select representative users, DPL employs an embedding model; in our implementation, we use the same embedding model as in our method.
    \item \textbf{PPlug}~\cite{pplug}: A plug-and-play personalization method that encodes a user's history into a dense user-specific embedding through a lightweight user embedder.
    This embedding is constructed via input-aware attention over user histories. The resulting embedding, along with an instruction embedding, is projected into the LLM input space via a trainable projector and prepended to the input to guide a frozen LLM.
    In our implementation of PPlug, we adopt the same user embedder as used in our proposed method.
\end{itemize}

\section{Implementation Details}\label{apd_implementation}

\subsection{Running Environments}

We implement all baseline methods and \ours with \texttt{Python 3.11.11}, \texttt{PyTorch}\footnote{\url{https://pytorch.org/}}~\cite{pytorch}, \texttt{transformers}\footnote{\url{https://huggingface.co/}}~\cite{transformers}, and \texttt{vLLM}\footnote{\url{https://github.com/vllm-project/vllm}}~\cite{vllm}.
To train the model, we utilize the \texttt{transformers} library.
Besides, we employ the \texttt{vLLM} library as the inference engine for both validation and testing, and adapt our model accordingly to ensure compatibility.

\subsection{Hyperparameter Configurations}

\subsubsection{Method Parameters}




In our implementation, the SAE model is implemented as a two-layer feed-forward network, consisting of an encoder that projects input embeddings from dimension $d=1024$ to a lower-dimensional latent space of size $d'=512$, and a decoder that reconstructs the input.
For the sparsity parameter $\rho$, we set it to 0.05.
To align the SAE output with the LLM input space, we employ two independent projection networks $\mathcal{M}_\text{his}$ and $\mathcal{M}_\text{diff}$, each implemented as a two-layer MLP with GELU activations, mapping the latent representation $z$ to the LLM embedding space.
Additionally, we use $\lambda = 100$ and $\gamma = 1\text{e}{-3}$ to balance the reconstruction and sparsity losses during training.

At most 8 user history entries are retrieved for each instance. If the input exceeds the context length limit, excess histories are discarded to ensure compatibility.

\subsubsection{Training Settings}

Before training, we initialize the model parameters using Xavier uniform initialization~\cite{xavier}.
We train the model using the \texttt{AdamW}~\cite{adamw} optimizer for a maximum of 8 epochs. 
The learning rate is set to 1e-5 with a weight decay of 0.025.
We apply a warmup ratio of 0.01 at the beginning of training.
The batch size per device is 1, and the gradient accumulation steps are 16 to achieve an effective batch size of 16.
We also enable \texttt{bfloat16} mixed precision and incorporate flash attention~\cite{flashattention2}.
Additionally, the training is conducted using DeepSpeed\footnote{\url{https://github.com/deepspeedai/DeepSpeed}}~\cite{zero,deepspeed} ZeRO Stage 1 optimization.

\subsubsection{Inference Settings}

We configure the model with a maximum length of 2048 tokens for both input and output. 
During inference for both validation and test, the temperature is set to 0.8, and the parameter top\_p is 0.95.

\section{Complete Ablation Studies \& Additional Experiments}\label{complete_ablation}

\begin{figure}[t]
    \centering
    \includegraphics[width=1\linewidth]{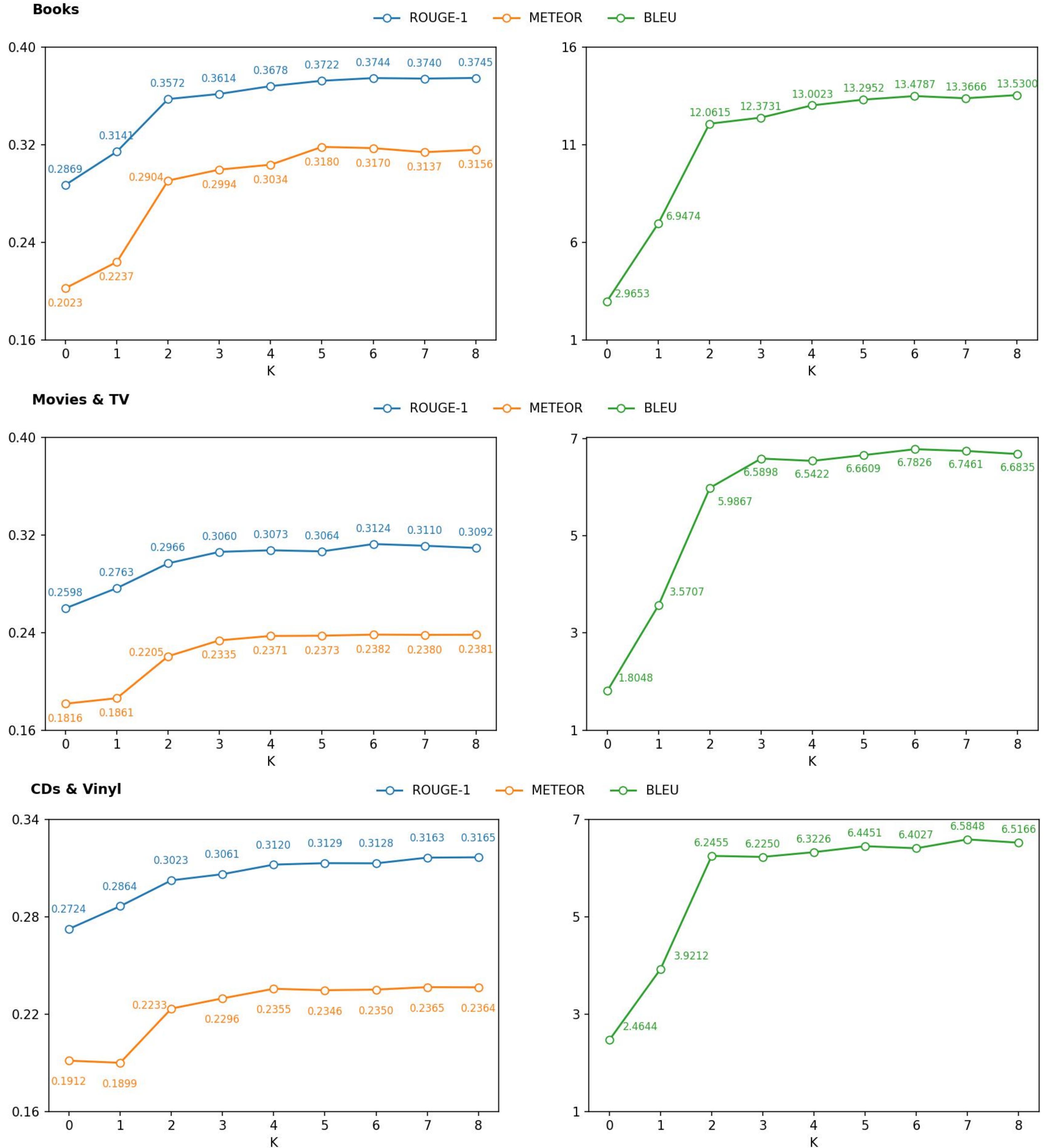}
    \caption{Detailed evaluation results across all three datasets (Books, Movies \& TV, CDs \& Vinyl) with varying numbers of retrieved user histories ($K$). The left figure shows ROUGE-1 and METEOR scores, and the right figure demonstrates BLEU scores.}
    \label{num_retrieve_complete}
\end{figure}

\setlength{\dashlinedash}{6pt}
\begin{table*}[!ht]
    \centering
    \small
    \caption{Complete ablation study on different configurations of user embeddings.}
    \renewcommand{\arraystretch}{1.2}
    \setlength{\tabcolsep}{4pt}
    \resizebox{0.96\textwidth}{!}{
    \begin{tabularx}{\textwidth}{>{\centering\arraybackslash}m{0.4cm} >{\centering\arraybackslash}m{2.6cm} *{9}{>{\centering\arraybackslash}m{1.1cm}}}
        \toprule
        \multicolumn{2}{c}{\textbf{Datasets ($\rightarrow$)}} & \multicolumn{3}{c}{\textbf{Books}} & \multicolumn{3}{c}{\textbf{Movies \& TV}} & \multicolumn{3}{c}{\textbf{CDs \& Vinyl}} \\
        \cmidrule(lr){3-5}
        \cmidrule(lr){6-8}
        \cmidrule(lr){9-11}
        \multicolumn{2}{c}{\textbf{Methods ($\downarrow$)}}  & R-1 & MET. & BL. & R-1 & MET. & BL. & R-1 & MET. & BL. \\
        \midrule
        \multicolumn{2}{c}{Non-Perso-7B} & 0.2907 & 0.1735 & 1.9766 & 0.2469 & 0.1503 & 0.7242 & 0.2604 & 0.1561 & 1.0997 \\
        \noalign{\vskip 2pt}
        \hdashline
        \noalign{\vskip 2pt}
        \multirow{3}{*}{\rotatebox{90}{\textit{w/o text}}} 
        & his\_emb & 0.2912 & 0.1718 & 2.4364 & 0.2545 & 0.1625 & 1.7048 & 0.2726 & 0.1711 & 2.1962 \\
        & diff\_emb & 0.3022 & 0.1839 & 2.6648 & 0.2542 & 0.1546 & 0.8574 & 0.2690 & 0.1616 & 1.2601 \\
        & his\_emb + diff\_emb & 0.2970 & 0.2227 & 5.5622 & 0.2586 & 0.1871 & 3.5629 & 0.2713 & 0.1853 & 3.3092 \\
        \noalign{\vskip 2pt}
        \hdashline
        \noalign{\vskip 2pt}
        \multirow{3}{*}{\rotatebox{90}{\textit{w/ text}}}
        & his\_emb & 0.3722 & 0.3110 & 12.9361 & 0.3026 & 0.2332 & 6.0120 & 0.3051 & 0.2268 & 5.3390 \\
        & diff\_emb & 0.3596 & 0.2781 & 10.6435 & 0.2964 & 0.2128 & 5.1985 & 0.3049 & 0.2108 & 4.9141 \\
        & his\_emb + diff\_emb (ours) & \textbf{0.3745} & \textbf{0.3156} & \textbf{13.5300} & \textbf{0.3092} & \textbf{0.2381} & \textbf{6.6835} & \textbf{0.3165} & \textbf{0.2364} & \textbf{6.5166} \\
        \bottomrule
    \end{tabularx}}
    \label{ablation_1}
\end{table*}
\begin{table*}[!t]
    \centering
    \caption{Complete ablation study on representation refinement.}
    \renewcommand{\arraystretch}{1.2}
    \resizebox{0.78\textwidth}{!}{
    \begin{tabularx}{\textwidth}{>{\centering\arraybackslash}m{2.4cm} *{9}{>{\centering\arraybackslash}m{1cm}}}
        \toprule
        \textbf{Datasets ($\rightarrow$)} & \multicolumn{3}{c}{\textbf{Books}} & \multicolumn{3}{c}{\textbf{Movies \& TV}} & \multicolumn{3}{c}{\textbf{CDs \& Vinyl}} \\
        \cmidrule(lr){2-4}
        \cmidrule(lr){5-7}
        \cmidrule(lr){8-10}
        \textbf{Methods ($\downarrow$)}  & R-1 & MET. & BL. & R-1 & MET. & BL. & R-1 & MET. & BL. \\
        \midrule
        \textit{w/o DR} & 0.3704 & 0.3016 & 13.3651 & 0.3091 & 0.2325 & 6.5149 & 0.3039 & 0.2283 & 5.6812 \\
        \textit{w/ AE} & 0.3691 & 0.2994 & 12.5453 & 0.3084 & 0.2350 & 6.5949 & 0.3167 & 0.2355 & 6.4352 \\
        \textit{w/ SAE} & \textbf{0.3745} & \textbf{0.3156} & \textbf{13.5300} & \textbf{0.3092} & \textbf{0.2381} & \textbf{6.6835} & \textbf{0.3165} & \textbf{0.2364} & \textbf{6.5166} \\
        \bottomrule
    \end{tabularx}}
    \label{ablation_2}
\end{table*}

\subsection{User Embedding Configuration}

In this section, we provide the complete results for different user embedding configurations evaluated in our ablation study.
While the main paper only reports METEOR scores in Table~\ref{ablation_meteor_1}, we include here the full results for all three metrics (ROUGE-1, METEOR, and BLEU) across all datasets.
The results in Table~\ref{ablation_1} offer a more comprehensive view of how different embedding types (\textit{his\_emb}, \textit{diff\_emb}) and the presence or absence of retrieved text affect personalization performance.

\begin{table*}[!t]
    \centering
    \caption{Performance comparison between different retrieval strategies across the three datasets.}
    \renewcommand{\arraystretch}{1.2}
    \resizebox{0.78\textwidth}{!}{
    \begin{tabularx}{\textwidth}{>{\centering\arraybackslash}m{2.4cm} *{9}{>{\centering\arraybackslash}m{1cm}}}
        \toprule
        \textbf{Datasets ($\rightarrow$)} & \multicolumn{3}{c}{\textbf{Books}} & \multicolumn{3}{c}{\textbf{Movies \& TV}} & \multicolumn{3}{c}{\textbf{CDs \& Vinyl}} \\
        \cmidrule(lr){2-4}
        \cmidrule(lr){5-7}
        \cmidrule(lr){8-10}
        \textbf{Methods ($\downarrow$)}  & R-1 & MET. & BL. & R-1 & MET. & BL. & R-1 & MET. & BL. \\
        \midrule
        Random & 0.3287 & 0.2573 & 5.4657 & 0.2955 & 0.2125 & 2.6946 & 0.3064 & 0.2138 & 2.9218 \\
        BM25 & \underline{0.3325} & \underline{0.2650} & \underline{5.9851} & 0.2953 & 0.2123 & \underline{2.7802} & 0.3066 & 0.2148 & 2.9832 \\
        Contriever & \underline{0.3325} & 0.2608 & 5.7479 & \underline{0.2958} & \underline{0.2128} & 2.7584 & \underline{0.3077} & \underline{0.2160} & \underline{3.0204} \\
        Recency & \textbf{0.3404} & \textbf{0.2735} & \textbf{6.8178} & \textbf{0.2983} & \textbf{0.2142} & \textbf{2.8680} & \textbf{0.3092} & \textbf{0.2177} & \textbf{3.1588} \\
        \bottomrule
    \end{tabularx}}
    \label{rag_methods}
\end{table*}
\begin{table*}[!t]
    \centering
    \caption{Performance comparison with and without system prompt guidance.}
    \renewcommand{\arraystretch}{1.2}
    \resizebox{0.78\textwidth}{!}{
    \begin{tabularx}{\textwidth}{>{\centering\arraybackslash}m{2.4cm} *{9}{>{\centering\arraybackslash}m{1cm}}}
        \toprule
        \textbf{Datasets ($\rightarrow$)} & \multicolumn{3}{c}{\textbf{Books}} & \multicolumn{3}{c}{\textbf{Movies \& TV}} & \multicolumn{3}{c}{\textbf{CDs \& Vinyl}} \\
        \cmidrule(lr){2-4}
        \cmidrule(lr){5-7}
        \cmidrule(lr){8-10}
        \textbf{Methods ($\downarrow$)}  & R-1 & MET. & BL. & R-1 & MET. & BL. & R-1 & MET. & BL. \\
        \midrule
        \textit{w/o} Guidance & 0.3704 & 0.3016 & 13.3651 & 0.3091 & 0.2325 & 6.5149 & 0.3039 & 0.2283 & 5.6812 \\
        \textit{w/} Guidance & \textbf{0.3745} & \textbf{0.3156} & \textbf{13.5300} & \textbf{0.3092} & \textbf{0.2381} & \textbf{6.6835} & \textbf{0.3165} & \textbf{0.2364} & \textbf{6.5166} \\
        \midrule
        \textit{+Improvement} & \textit{0.0041} & \textit{0.0140} & \textit{0.1649} & \textit{0.0001} & \textit{0.0056} & \textit{0.1686} & \textit{0.0126} & \textit{0.0081} & \textit{0.8354} \\
        \bottomrule
    \end{tabularx}}
    \label{guidance}
\end{table*}

\subsection{Representation Refinement}

This section presents the complete results for the different representation refinement strategies discussed in our ablation study.
Table~\ref{ablation_2} reports ROUGE-1, METEOR, and BLEU scores for the \textit{w/o DR}, \textit{w/ AE}, and \textit{w/ SAE} settings across all datasets, providing a more detailed understanding of their relative effectiveness.

\subsection{Impact of History Number}\label{history_number}

We provide the full results across all evaluation metrics in Figure~\ref{num_retrieve_complete}.
As shown in the figure, all three evaluation metrics (ROUGE-1, METEOR, and BLEU) exhibit a consistent upward trend across the three datasets as the number of retrieved histories ($K$) increases.
This improvement can be attributed to the additional contextual information provided by retrieved histories, along with our injected user-specific embedding and difference-aware embedding.
Notably, the most significant gains occur when $K$ increases from 0 to 3, especially for the BLEU metric.
Beyond this range, the performance tends to plateau, with only marginal improvements or slight fluctuations.
A slight dip is observed in METEOR on the \textit{CDs \& Vinyl} dataset when $K$ increases from 0 to 1, which may result from noise or limited informativeness in the single retrieved history.
As more histories are incorporated, the signal becomes more stable and representative, leading to consistent improvements.

Overall, these results demonstrate that our method substantially enhances the RAG pipeline.
The retrieve-and-inject paradigm we adopt proves to be a strong and effective framework for personalization.


\subsection{Retrieval Method}

To investigate the impact of different retrieval strategies and identify the most effective one for use in both the baselines and our method, we evaluate four retrieval approaches: random, BM25~\cite{bm25}, Contriever~\cite{contriever}, and recency (the most recent).
Experiments are conducted using the \texttt{Qwen2.5-32B-Instruct} model, and the results are presented in Table~\ref{rag_methods}.

As shown in Table~\ref{rag_methods}, the choice of retrieval strategy has a notable impact on generation performance.
The random retrieval baseline yields the lowest performance, indicating the importance of relevant context in guiding generation.
BM25 and Contriever perform comparably, with slight advantages in different metrics.
Among the four methods evaluated, the recency-based retrieval consistently outperforms the others across all metrics.
Based on these results, we adopt the recency retrieval strategy in all subsequent experiments.

\subsection{System Prompt Guidance}\label{system_prompt_guidance}

As shown in Figure~\ref{prompt}, we incorporate additional information into the system prompt to help the model better understand the injected personalization prompts.
To assess its effectiveness, we conduct experiments to analyze the impact of this guidance.
Table~\ref{guidance} reports the results across all datasets and evaluation metrics.
We observe that incorporating system prompt guidance consistently improves performance across the board.
Hence, we adopt the system prompt guidance by default in all experiments.

\section{Further In-Depth Analysis}

\subsection{Interpretability Analysis}

To further examine how personalization is achieved, we conducted an interpretability analysis comparing DEP with the baseline DPL. The results show that DEP better captures users’ word choice, semantic information, and overall writing patterns, thereby aligning generated reviews more closely with users’ authentic writing style.

Table~\ref{interpretability} presents an illustrative example. Compared with DPL, which produces a more formal and detached review, DEP generates text that reflects the user’s actual phrasing, tone, and evaluative stance. This demonstrates DEP’s ability to internalize and reproduce the personalized linguistic patterns of users.

\begin{table}[t]
\centering
\small
\resizebox{0.48\textwidth}{!}{
\begin{tabular}{p{0.15\linewidth} p{0.85\linewidth}}
\toprule
\textbf{Source} & \textbf{Review Text} \\
\midrule
DPL-generated Review & On the positive side, the film does a good job of maintaining a level of respect for its audience by not overloading with graphic gore, which is a relief for those who might find that kind of content too disturbing. \\
\addlinespace
\midrule
DEP-generated Review & I'll give it that, it is not full of gore and disgust. If you are looking for a good horror film, this is not it. \\
\addlinespace
\midrule
User's Actual Review & Not something I'm eager to watch again, but in a pinch, it is at least not full of gore and disgust. I'll give it that. \\
\bottomrule
\end{tabular}
}
\caption{Comparison between reviews generated by DPL and DEP and the user’s actual review.}
\label{interpretability}
\end{table}

\subsection{Practical Applicability}

To assess the practical applicability of our method, we further examine its training cost. Importantly, our approach does not involve tuning the LLM itself; instead, the backbone model remains fixed while only the input component is tuned. This design substantially reduces the number of trainable parameters to approximately 0.4\% of the LLM’s total parameters. Consequently, the additional computational overhead is minimal. In practice, the training cost is comparable to the soft prompt tuning baseline PPlug, with both methods requiring around 40 minutes per epoch on our datasets using a single GPU.

\section{Overview of Templates \& Prompts}\label{ov_prompt}

\begin{figure}[t]
    \centering
    \includegraphics[width=1\linewidth]{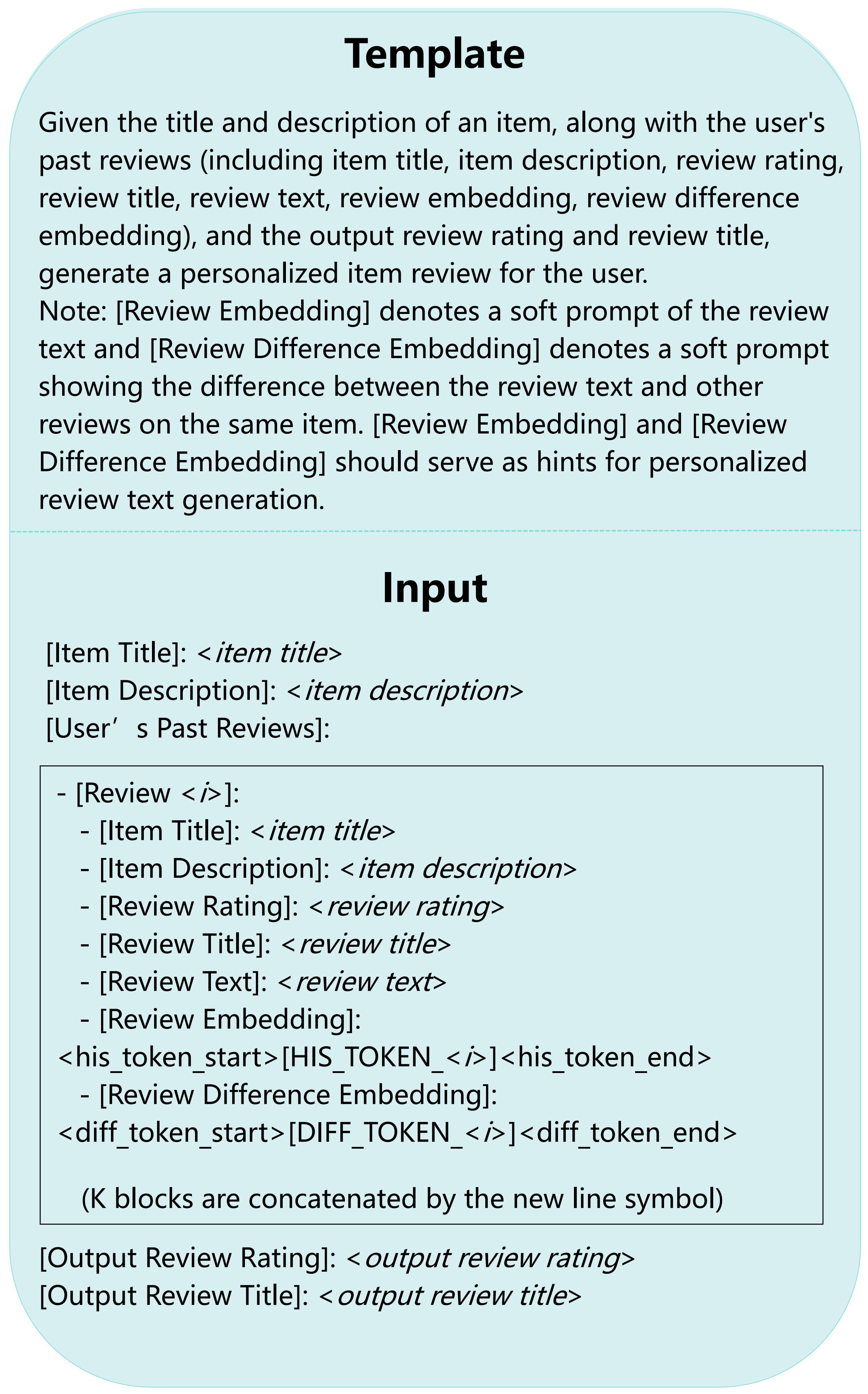}
    \caption{The system prompt template and input template for the base LLM.}
    \label{prompt}
\end{figure}

In this section, we illustrate the prompt design used in our framework.
As shown in Figure~\ref{prompt}, the upper part depicts the system prompt, which defines the model's global behavior and task instruction.
The lower part shows an example of the input prompt, including retrieved user histories and object descriptions, which are fed into the model for generation.
This prompt structure follows the retrieve-and-inject paradigm, where both user-specific and difference-aware embeddings are embedded via soft prompts \textit{[HIS\_TOKEN\_i]} and \textit{[DIFF\_TOKEN\_i]} to guide the generation.
The four special tokens \textit{<his\_token\_start>}, \textit{<his\_token\_end>}, \textit{<diff\_token\_start>}, and \textit{<diff\_token\_end>} are introduced to explicitly mark the boundaries of user-specific and difference-aware embeddings in the input sequence.
The note part is the system prompt guidance described in Section~\ref{system_prompt_guidance}.

\begin{figure*}[t]
    \centering
    \includegraphics[width=0.9\linewidth]{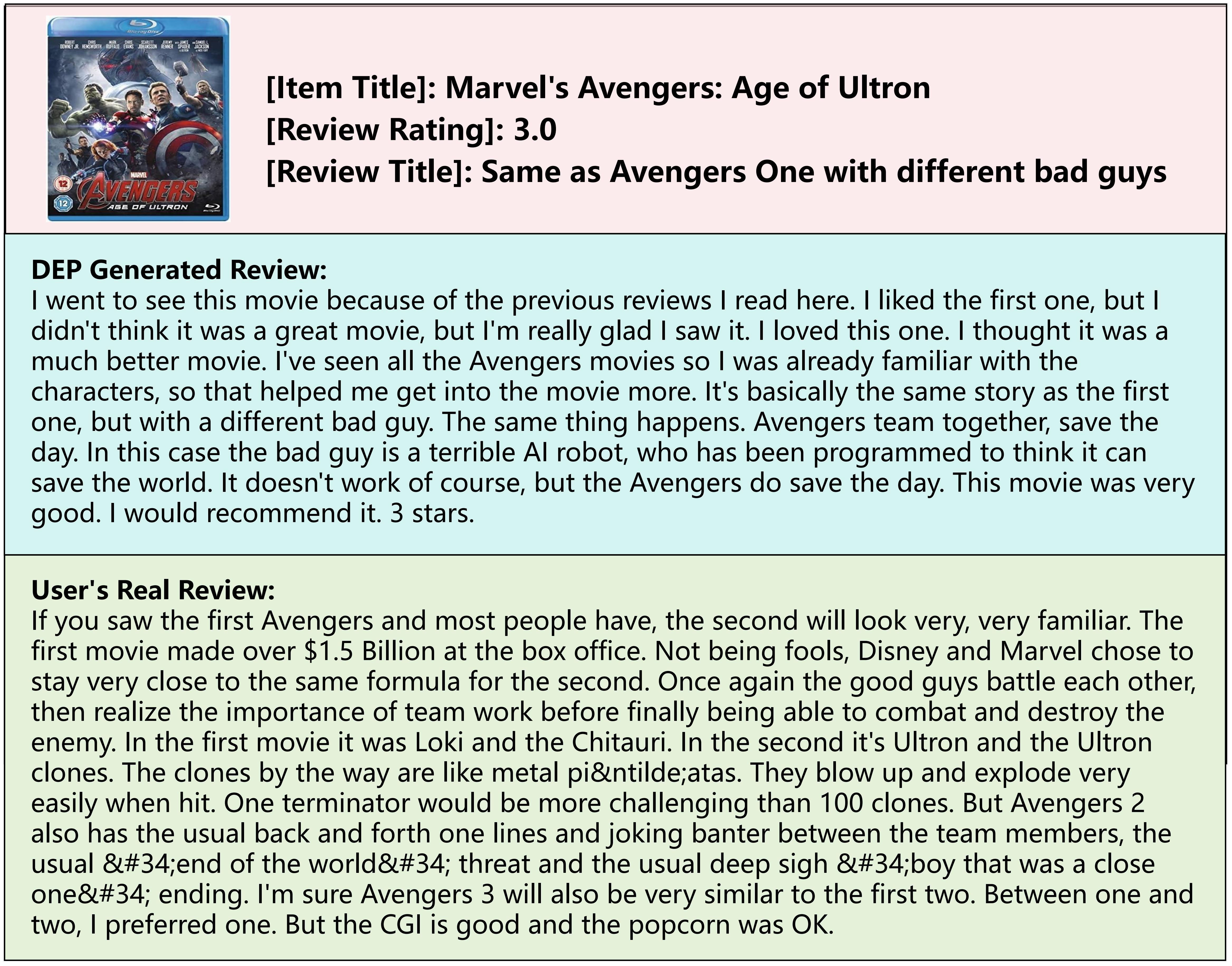}
    \caption{A case study which compares the \textsc{DEP}-generated review and the user's real review for the item movie \\ \textit{Avengers: Age of Ultron}.}
    \label{case_study}
\end{figure*}

\section{Case Study}\label{apd_case_studies}

In this section, we present a case study to illustrate the output generated by our framework as shown in Figure~\ref{case_study}.

In this example, the review generated by \ours closely aligns with the user's real review in both content and sentiment.
Both reviews highlight the central observation that \textit{Avengers: Age of Ultron} feels very similar to the first Avengers movie, with the main difference being the villain.
Moreover, \ours incorporates additional signals such as the user’s familiarity with the franchise and a moderately positive tone that matches the provided 3-star rating.
This case demonstrates that \ours can generate reviews that are not only coherent but also well-aligned with the user's original opinion, supporting the effectiveness of difference-aware modeling in the embedding space for personalization.

\end{document}